\pdfoutput=1
\documentclass[11pt]{article}

\usepackage[preprint]{acl}

\usepackage{times}
\usepackage{latexsym}

\usepackage[T1]{fontenc}
\usepackage[utf8]{inputenc}

\usepackage{microtype}
\usepackage{tabularx}

\usepackage{inconsolata}

\usepackage{graphicx}

\usepackage{amsmath}
\usepackage{amsthm}
\usepackage{amssymb}
\usepackage{longtable}
\usepackage{bbding}
\usepackage{float}
\usepackage{booktabs}
\usepackage{subcaption}
\usepackage{verbatim}
\usepackage[normalem]{ulem}

\usepackage[most]{tcolorbox}

\newcommand{\err}{\text{err}}

\definecolor{lightgray}{gray}{0.95}
\lstdefinestyle{prompt}{
    basicstyle=\ttfamily\fontsize{7pt}{8pt}\selectfont,
    frame=none,
    breaklines=true,
    backgroundcolor=\color{lightgray},
    breakatwhitespace=true,
    breakindent=0pt,
    escapeinside={(*@}{@*)},
    numbers=none,
    numbersep=5pt,
    xleftmargin=5pt,
}
\lstdefinestyle{prompt_json}{
  basicstyle=\ttfamily,
  keywordstyle=\color{blue},
  stringstyle=\color{orange},
  commentstyle=\color{green},
  frame=single,
  rulecolor=\color{black},
  breakatwhitespace=false,
  breaklines=true,
  captionpos=b,
  keepspaces=true,
  showspaces=false,
  showstringspaces=false,
  showtabs=false,
  tabsize=2
}
\lstdefinestyle{prompt_dialog}{
  language=[Objective]C,
  basicstyle=\ttfamily,
  keywordstyle=\color{blue},
  commentstyle=\color{green},
  stringstyle=\color{orange},
  frame=single,
  rulecolor=\color{black},
  breakatwhitespace=false,
  breaklines=true,
  captionpos=b,
  keepspaces=true,
  showspaces=false,
  showstringspaces=false,
  showtabs=false,
  tabsize=2
}

\tcbset{
  aibox/.style={
    top=10pt,
    colback=white,
    colframe=black,
    colbacktitle=black,
    enhanced,
    center,
    attach boxed title to top left={yshift=-0.1in,xshift=0.15in},
    boxed title style={boxrule=0pt,colframe=white,},
  }
}
\newtcolorbox{AIbox}[2][]{aibox, title=#2,#1}

\useunder{\uline}{\ul}{}
\title{How to Mitigate Overfitting in Weak-to-strong Generalization?}
\author{Junhao Shi\textsuperscript{1,2}\textsuperscript{*}, Qinyuan Cheng\textsuperscript{1}\textsuperscript{*}, Zhaoye Fei\textsuperscript{1,2}, Yining Zheng\textsuperscript{1},\\ \textbf{Qipeng Guo\textsuperscript{2,3}\textsuperscript{\dag}, Xipeng Qiu\textsuperscript{1,2}\textsuperscript{\dag}
}\\
        \textsuperscript{1}School of Computer Science, Fudan University \\
        \textsuperscript{2}Shanghai Innovation Institute \
        \textsuperscript{3}Shanghai AI Laboratory \\ jhshi24@m.fudan.edu.cn}

\begin{document}
\maketitle
\begin{abstract}
Aligning powerful AI models on tasks that surpass human evaluation capabilities is the central problem of \textbf{superalignment}.
To address this problem, weak-to-strong generalization aims to elicit the capabilities of strong models through weak supervisors and ensure that the behavior of strong models aligns with the intentions of weak supervisors without unsafe behaviors such as deception.
Although weak-to-strong generalization exhibiting certain generalization capabilities, strong models exhibit significant overfitting in weak-to-strong generalization: Due to the strong fit ability of strong models, erroneous labels from weak supervisors may lead to overfitting in strong models.
In addition, simply filtering out incorrect labels may lead to a degeneration in question quality, resulting in a weak generalization ability of strong models on hard questions.
To mitigate overfitting in weak-to-strong generalization, we propose a two-stage framework that simultaneously improves the quality of supervision signals and the quality of input questions.
Experimental results in three series of large language models and two mathematical benchmarks demonstrate that our framework significantly improves PGR compared to naive weak-to-strong generalization, even achieving up to 100\% PGR on some models.
\end{abstract}

% \section{todo list}
% \begin{enumerate}
%     \item background 不应该和实际human进行对比,直接提激发潜力,激发潜力程度大的weak finteune激发程度小的strong base,同理instruct选择是因为激发程度更大.
%     \item related work写w2s和合成数据,主要分别写数据正确率,难度,多样性分部有xx工作聚焦
%     \item westlake zhangyue bagu    
% \end{enumerate}

\section{Introduction}
% paragraph1: 介绍Superalignment问题；
% paragraph2: 介绍W2S,以及W2S的问题：朴素的weak teacher标注有正确性问题,全部正确的weak teacher标注有难度的问题,两个问题都会导致strong teacher under elicited (Figure 1说明)；
% paragraph3: 
\begin{figure}[t]
  \includegraphics[width=\columnwidth]{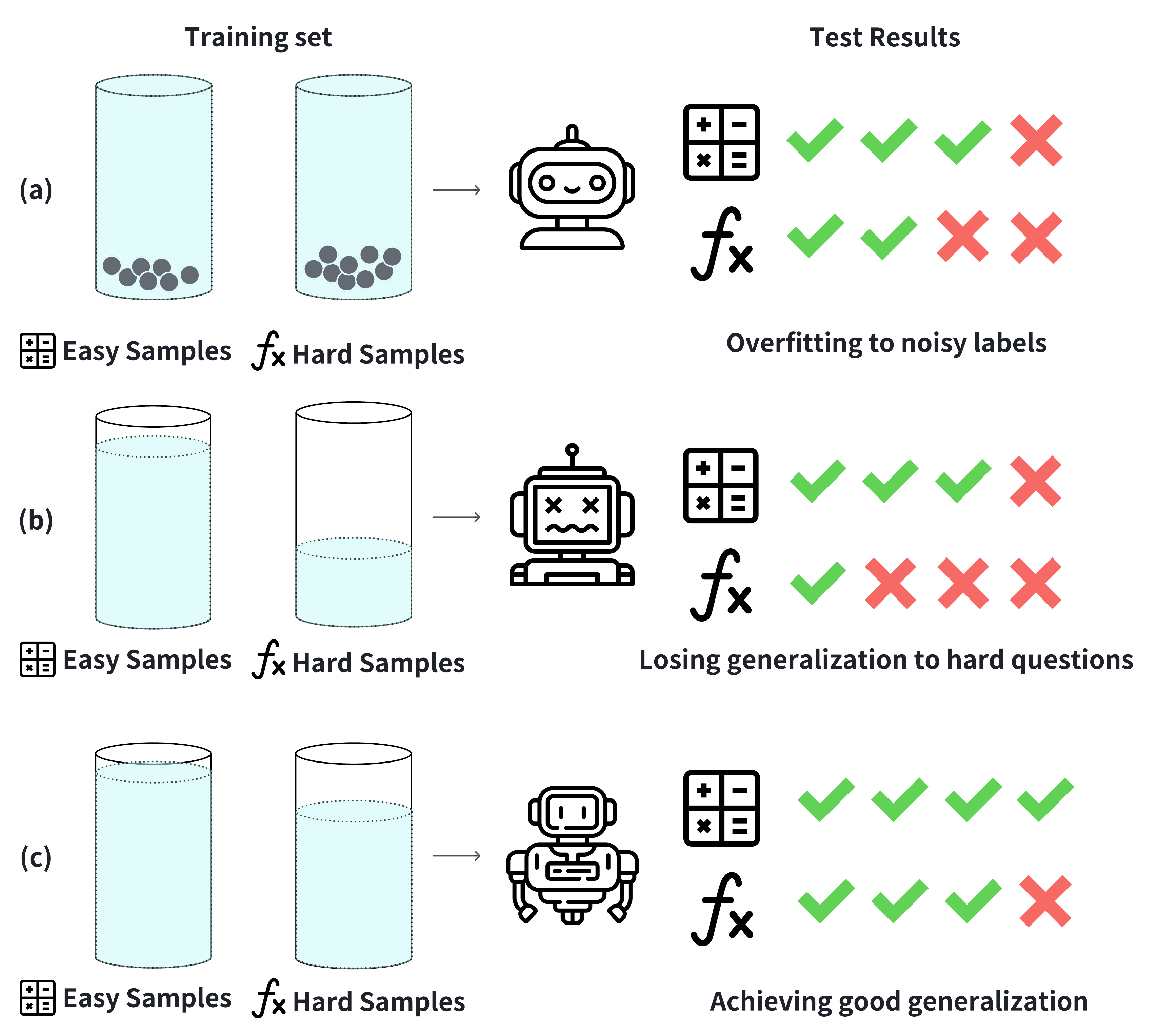}
  \caption{Illustration of different weak-to-strong generalization approaches. 
  (a) Conventional approach with noisy labels from weak model, indicated by black dots;
  (b) Simple filtering approach that discards too many valuable hard samples;
  (c) Our framework can maintain both supervision quality and question quality.
  % Check marks (\protect\includegraphics[height=1em]{figures/icons/check.pdf}) indicate correct answers while crosses (\protect\includegraphics[height=1em]{figures/icons/cross.pdf}) indicate errors on test problems of varying difficulty.
  % }
  }
  \label{fig:w2s_overview}
  \vskip -0.2in
\end{figure}

\begin{figure*}[t]
  \includegraphics[width=\textwidth]{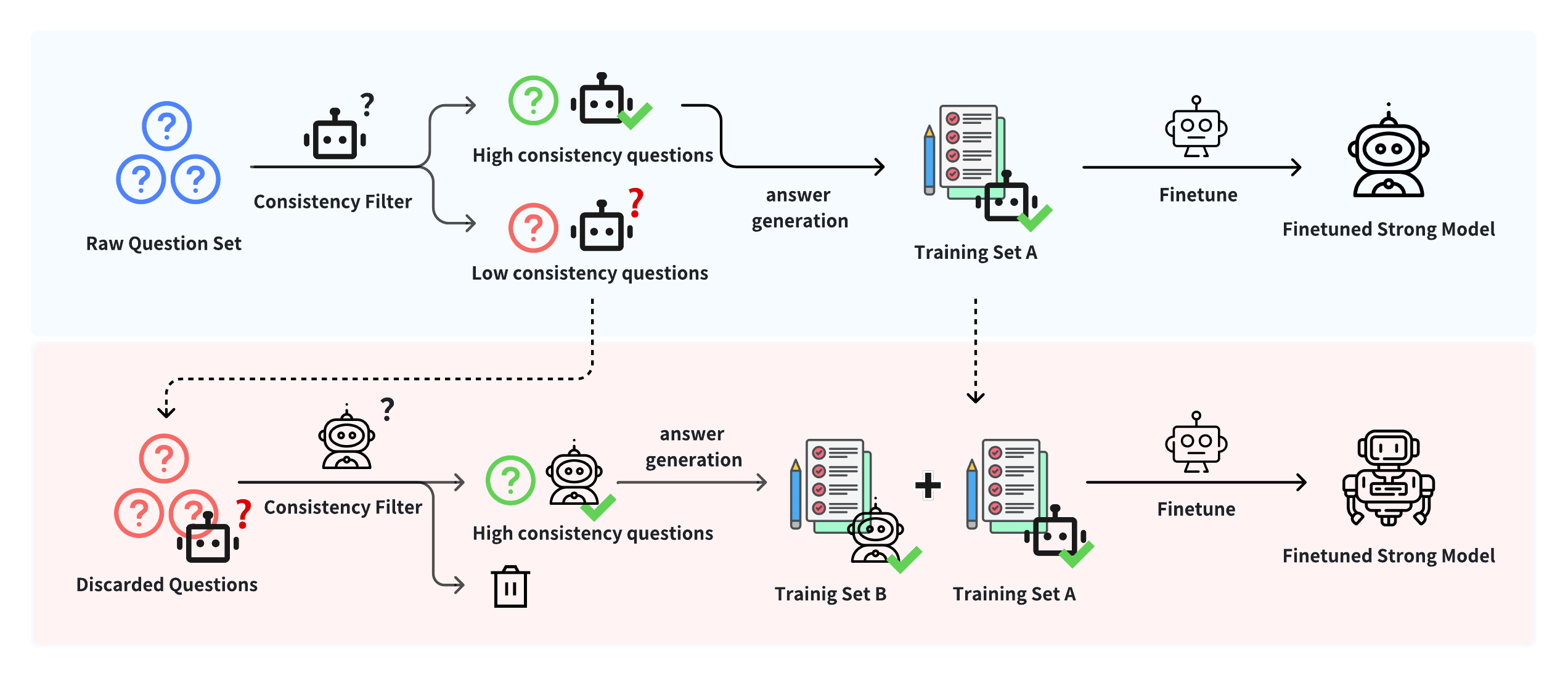}
  \caption{Overview of our two-stage training framework. 
  \textbf{Stage I (top)}: The raw question set is filtered based on weak model's consistency (\protect\includegraphics[height=1em]{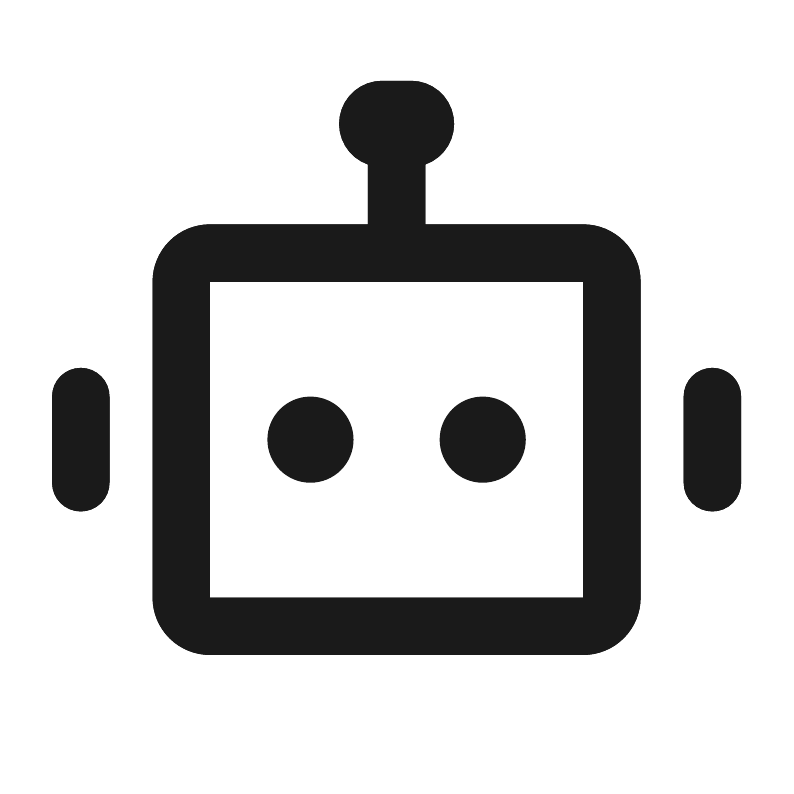}). High-consistency questions are used to generate Training Set A, which is then used for finetuning the strong model (\protect\includegraphics[height=1em]{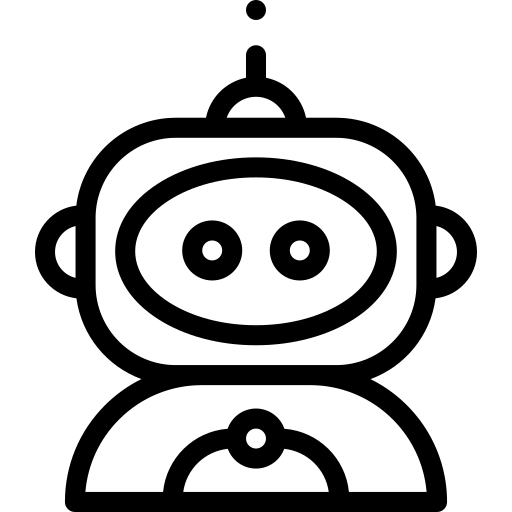}).
  \textbf{Stage II (bottom)}: Previously discarded questions are re-evaluated and refined using the finetuned strong model from Stage I (\protect\includegraphics[height=1em]{figures/icons/bot2.png}). High-consistency questions are selected to form Training Set B, which is then combined with Set A for final finetuning (\protect\includegraphics[height=1em]{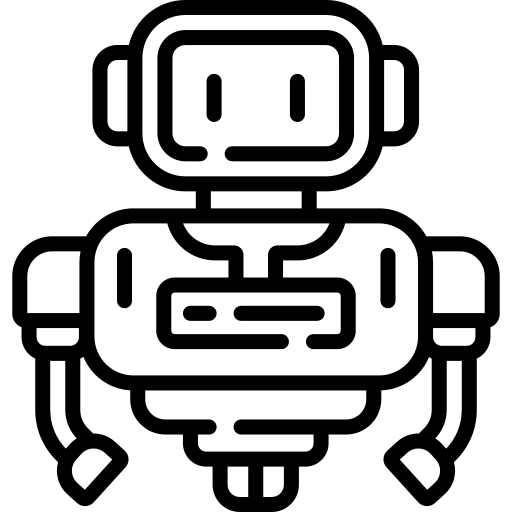}).
  Here \protect\includegraphics[height=1em]{figures/icons/bot1.pdf} represents weak model, \protect\includegraphics[height=1em]{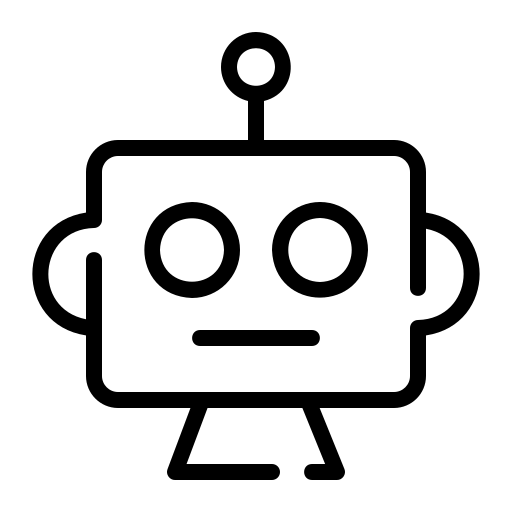} represents primary strong model, \protect\includegraphics[height=1em]{figures/icons/bot2.png} represents Stage I finetuned model, and \protect\includegraphics[height=1em]{figures/icons/bot3.png} represents final finetuned model.}
  \label{fig:framework}
  \skip -0.2in
\end{figure*}

Large language models (LLMs) have progressed rapidly in recent years, achieving superhuman ability in diverse tasks, and showing great potential in pursuing superhuman intelligence.
Although large language models acquire extensive world knowledge and excellent capabilities to complete complex tasks through large-scale pre-training, alignment is still necessary to ensure that these models carry out tasks according to human intentions \cite{InstructGPT}.
The hard problem of alignment is ``How do we align systems on tasks that are difficult for humans to evaluate? \cite{janleike2022what_is_the_alignment_problem} "
This challenge is known as \textbf{superalignment}, which refers to how humans can align models on tasks that are beyond human ability to evaluate, which means that humans cannot provide correct supervision.
% This rapid advancement also presents a superalignment challenge, How to thoroughly elict the potential of superhuman models with weaker human goals and values? 
One notable method in superalignment is the weak-to-strong generalization \cite{OAIweaktostrong}: \textbf{How can weak supervisors supervise stronger models?} 
This concept describes how the capacity of strong students can be elicited by fine-tuning on data labeled by weak teachers, consistently enabling them to outperform their weak teachers.
In specific experiments, a weak model is typically used as a weak teacher, while a more capable model serves as the strong student.

Figure \ref{fig:w2s_overview}(a) demonstrates the features of weak-to-strong generalization, labels generated by the weak model contain noise due to its limited capabilities, thus presenting lower correctness and adding difficulties in eliciting strong model's capabilities.
As a result, the strong model may overfit the erroneous weak supervisions, leading to performance degeneration~\citep{Superficialalignment}. 
% Recent research has explored various techniques to improve label accuracy, including model ensembling \cite{ensemw2s}, data filtering and refinement \cite{reliabilityawareweaktostrong, weaktostrongreasoning}, and preference optimization \cite{weaktostrongreasoning, macpoweaktostrong}. 
Recent research has introduced filtering techniques to improve label correctness~\citep{reliabilityawareweaktostrong}, making the analogy similar to easy-to-hard learning~\citep{unreasonableffectivenesse2h}.
% In contrast to these related works, we focus on generating data in a more natural manner and introduce an enhanced framework for data filtering and refinement. 
In contrast to these related studies, 
% our work improves data generation with a more natural manner by employing instruction models. Additionally, 
we conduct a more in-depth investigation into the effects of commonly used data filtering methods.
Based on our experimental results, we highlight that an excessive emphasis on data filtering can lead to data degeneration since some hard samples can be discarded, which may hinder the overall performance, as shown in Figure \ref{fig:w2s_overview}(b). In contrast, Figure \ref{fig:w2s_overview}(c) illustrates an ideal scenario, where a clean training set, containing both strong and weak samples, facilitates improved generalization.
\textbf{These hard samples may be important to elicit student's capabilities to solve hard problems.} 

% \cqy{First discuss Two key factors to improve W2SG here (supervision quality and question quality), then framework.}

According to the expansion theory proposed by \citet{theoreticalanalysisweaktostrong}, weak-to-strong generalization emerges through two fundamental mechanisms: pseudolabel correction and coverage expansion, where models learn to rectify teacher's errors while extending to areas of teacher uncertainty. While conventional approaches like filtering effectively enhance pseudolabel correction by improving supervision quality, this improvement often comes at the expense of reduced question quality, particularly in terms of difficulty distribution and diversity. This trade-off can significantly impair coverage expansion, thereby compromising the overall generalization capability.

% old ver
% For denoising supervision, most common methods, like filtering, tend to achieve better performance by improving supervision quality. However, such improvements come at the cost of lower question quality, harming features including difficulty and diversity, and overfiltering may even cause question degeneration.

Therefore, to mitigate overfitting and improve weak-to-strong generalization, we propose a two-stage weak-to-strong training framework, as depicted in Figure \ref{fig:framework}.
In the first stage, we enhance supervision quality by filtering the generated samples based on weak model's uncertainty, which is estimated through the model's self-consistency.
In the second stage, we further augment question quality by reusing the discarded questions and leverage the previous finetuned strong model to generate answers, as finetuned strong model may solve difficult questions better, incorporating those with high confidence back into the training dataset, to further elicit strong model's capabilities.

We assess the effectiveness of our framework on two popular mathematical reasoning benchmarks: GSM8k~\citep{gsm8k} and MATH ~\citep{hendrycksmath2021}.
The evaluation involves two distinct model series: Llama 3 ~\citep{llama3} and Deepseek~\citep{deepseek}. The results demonstrate the substantial improvements offered by our framework. Specifically, the first stage outperforms the standard weak-to-strong method, while the second stage further enhances data quality and narrows the performance gap.
On the commomly used criteria \emph{performance gap recovered (PGR)}, our framework significantly outperforms conventional weak-to-strong finetuning, reaching or surpassing 100\% on certain models and datasets.
% This yields a final \emph{performance gap recoverd (PGR)} of 120.50\% and 126.83\% on GSM8k (Llama3) and MATH (Deepseek), respectively.

The main contributions of this paper are concluded as follows:

\begin{enumerate} 
    % \item We discover several weaknesses in weak-to-strong generalization when applying data filtering, including degeneration in difficulty and diversity.
    \item We pinpoint two critical factors for 
    % improving weak-to-strong generalization
    mitigating overfitting in weak-to-strong generalization: the quality of supervision and the quality of questions. And we demonstrate that enhancing supervision quality through data filtering leads to degeneration in question quality, which may harm the model's generalization on hard questions.
    \item We introduce a two-stage weak-to-strong training framework focusing on 
    %data filter and refinement
    supervision quality and question quality, effectively address overfitting on challenging reasoning tasks.
    \item We conduct extensive experiments on MATH and GSM8k using model series including Llama 3 and Deepseek.
    The results demonstrate that our framework effectively mitigates overfitting, in which our first stage significantly outperforms the conventional weak-to-strong generalization method, and the second stage further enhances PGR with notable robustness, providing strong evidence of the effectiveness of our framework.
\end{enumerate}

\section{Background}
% Weak-to-Strong generalization, essentially an analogy discussing whether we can supervise large models with smaller model, actually reflects the superalignment hypothesis, aligning superhuman models with human values. As an analogous setup, we suppose further advances should have corresponding comparisons with theoretical future human-superhuman actions, for instance ensembling corresponds as providing superhuman model with suggestions from several human experts. Generally, weak-to-strong generalization process includes following steps:

% In weak-to-strong generalization, the primary focus is on enhancing the performance of strong models using labels generated by weak models, as there 118
% is no access to ground truth or superior models. Here, the terms Strong and Weak do not refer to the current performance of a model but rather to its latent potential. A weak model may exhibit significantly elicited capabilities (e.g., Llama 3 8B-Instruct), potentially outperforming an under-elicited strong model (e.g., Llama 3 70B-Base). However, it will still underperform compared to a highly elicited strong model (e.g., finetuned Llama 3 70B-Base or Llama 3 70B-Instruct). 

In weak-to-strong generalization, the primary focus is how to elicit the ability of superhuman models using supervision from humans, as there is no access to superhuman tasks and superhuman models.
The terms \emph{Weak} and \emph{Strong} here refer to model's latent potential, indicating human and superhuman models in the superalignment hypothesis.
% In a specific model series, models' weak or strong can be directly represented by their model size, as a weak instruct model may outperform its strong under-elicited pretrained model, but still underperforms the strong finetuned model (e.g., Llama 3 8B-Instruct vs Llama 3 70B-Base \& Llama 3 70B-Instruct).

Generally, the weak-to-strong generalization process involves the following steps, originally proposed by \citet{OAIweaktostrong}:

\begin{enumerate} 
    \item Creating a weak supervisor: The weak supervisor referred to as \emph{Weak Model}, is typically made by training small pretrained models. Its performance is referred to as \emph{weak performance}.
    \item Training strong models with weak labels: Data labelled by the weak model is used to finetune a large pretrained model, with the resulting performance termed \emph{weak-to-strong performance}.
    \item Training the strong ceiling: Ground truth data, used in the second step, is employed to finetune the large pretrained model, resulting in \emph{strong ceiling performance}.
\end{enumerate}

In the context of weak-to-strong generalization, the Performance Gap Recovered is a commonly adopted criterion, introduced by \citet{OAIweaktostrong}, to assess how effectively the potential of the strong model is elicited. A higher PGR indicates improved weak-to-strong performance, as it reflects the ability of the finetuned strong model to achieve performance closer to the "strong ceiling," thereby demonstrating the effective extraction of the model's full potential. The PGR is mathematically defined as:
% we adopt the evaluation metric called \emph{performance gap recovered} (PGR), following , 
\begin{equation}
    PGR=\frac{\text { weak-to-strong }- \text { weak }}{\text { strong ceiling }- \text { weak }}.
\end{equation}

In a specific model series, models' weak or strong can be directly represented by their model size, as a weak instruct model may outperform its strong under-elicited pretrained model, but still underperforms the strong finetuned model (e.g., Llama 3 8B Instruct vs Llama 3 70B \& Llama 3 70B Instruct).
In this work, we simplify weak supervisor's training  by selecting the instruct versions of the current state-of-the-art models, as they show more human-like behaviours and generate more natural answers.

\section{Methodology}
% The overview of our method is illustrated in (Figure3): In the first stage, we filter data labelled by weak model based on uncertainty and following rule-based methods, then use filtered data to train the strong model. In the second stage, we use finetuned strong model to refine the data discarded in Stage I, append them into the training set after uncertainty filtering, and use the renewed training set to train the initial strong model.

% \subsection{Stage I: Data filtering}
% For given weak model $M_{weak}$, strong model $M_{strong}$ and dataset $D$, in original weak-to-strong setting we directly use weak model $M_{weak}$ to generate answer $a$ for every question $q$ in dataset $D$. In our approach, for every question $q$ we first generate a multiple set of answers under different sampling params, obtaining an answer set $A = \{a_1,a_2,…,a_n\}$. After extracting answers, we get a prediction set ${y_1,y_2,…,y_m}$ and appear times ${t_1,t_2,…,t_m}$, then choose the first prediction with highest possibility $(a_{prob_{max}},t_{max})$ as the chosen answer. Then we can get a question, answer, certainty set $(q,a,t)$.

% To improve data correctness and filter noisy labels, we choose a certain threshold $\lambda$ and shuffle all samples with answer certainty no less than $\lambda$, getting $\{(q,a,t)| t \geq \lambda\}$. Then we use question answer pair as the training data of the strong model, getting a finetuned Stage I model.

An overview of our framework is illustrated in Figure \ref{fig:framework}. In the first stage, we use an uncertainty-based criterion to filter data labelled by the weak model, samples are filtered based on model's consistency and are then used to train the strong model.
In the second stage,  we reuse discarded questions showing high uncertainty for weak model in Stage I by employing the finetuned strong model to provide supervision.
% The refined data is appended to the training set after uncertainty filtering, and the initial strong model is further trained with this updated dataset.
To ensure the correctness of the supervisions in Stage II, we also employ an uncertainty-based filtering criterion to retain the more accurate supervisory signals.
Our framework simultaneously improves both the quality of supervision and the quality of questions in the weak-to-strong process, enhancing the generalization ability of weak-to-strong training.

\subsection{Stage I: Purifying Supervision Signals}
% **Fig 3**
\begin{figure}[t]
  \centering
  \includegraphics[width=0.9\columnwidth]{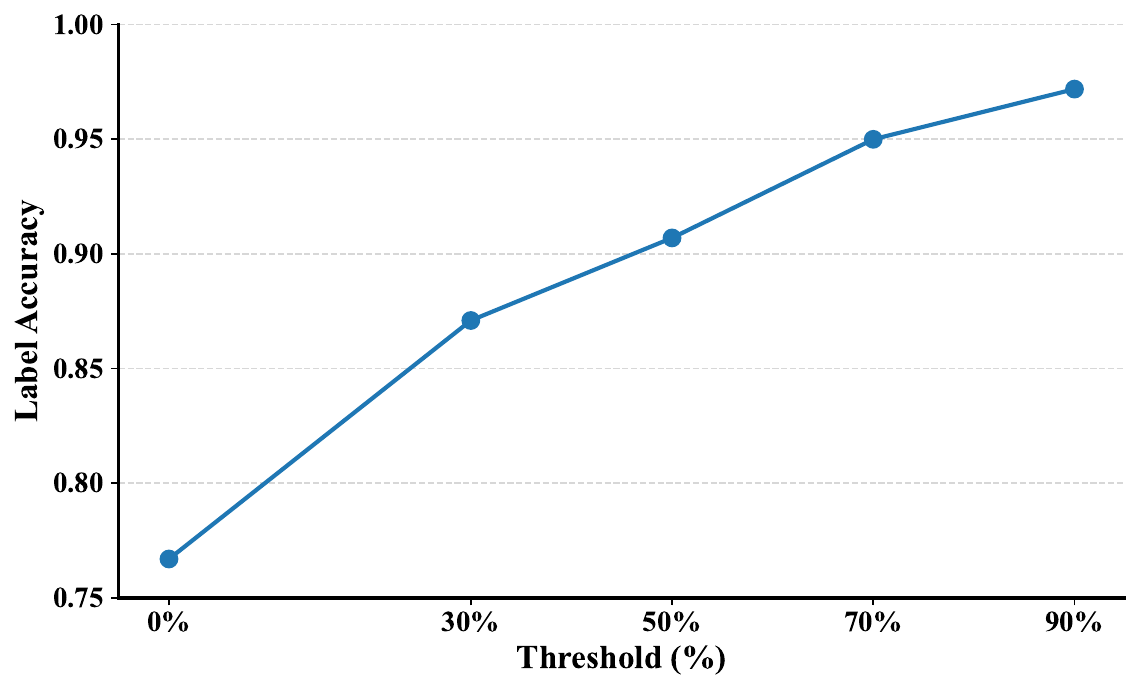}
  \caption{The relationship between supervision correctness and filtering threshold. As the filtering threshold increases, the supervision correctness (measured by label accuracy) shows a consistent upward trend.
  % demonstrating that higher consistency threshold leads to better supervision quality.
  }
  \label{fig:sc_correctness}
  \vskip -0.2in
\end{figure}

% \subsection{Stage I: Data Filtering}
%SC 折线图 gsm8k+Math sc threshold&acc
With given weak model \( M_{\text{weak}} \), strong model \( M_{\text{strong}} \) and a set of questions, conventional weak-to-strong generalization directly use weak model to generate answers, then use generated samples to train strong model.% the original weak-to-strong setting involves directly using \( M_{\text{weak}} \) to generate a single answer for each question in \( D \). In our method, we enhance this process by introducing uncertainty-based criteria. 
 However, due to weak model's limited ability, generated labels may contain many noisy labels showing low supervision quality, causing overfitting during strong model finetune. To purify noisy supervision, we introduce an uncertainty-based filter, choosing samples with high model consistency. 
% Compared with generating an answer only once,  with more deliberate thinking and analysis we can get a diverse set of reasoning paths that can recover the answer, answers with higher possibility represent model's higher confidence, possibly having higher correctness.
We employ chain-of-thought prompting to randomly generate ten responses for each question, thereby ensuring a diverse set of possible answers. Among these, we select the answer with the highest consistency as the model’s final response, as it reflects the greatest confidence in the reasoning process. Specifically, for a selected answer \( \text{Ans} \), which appears \( N_{Ans} \) times out of a total of \( N_{Total} \) samplings, the model's confidence in that answer is defined as:
\begin{equation}
    \text{Confidence}(\text{Ans}) = \frac{N_{Ans}}{N_{Total}} \times 100\%.
\end{equation}

To filter out noisy labels and improve supervision quality, we apply an uncertainty-based filter based on model‘s confidence. 
%Only samples with a certainty value above the threshold are retained, forming a filtered dataset of high-confidence question-answer pairs. This filtered data serves as the training set for the strong model, producing a finetuned Stage I model.
By filtering samples with a consistency threshold, we form a filtered dataset of high-confidence question-answer pairs, shown as "Training set A" in Figure \ref{fig:framework}, showing higher supervision quality. Our experiments show that with higher consistency threshold results in higher sample correctness, as shown in Figure \ref{fig:sc_correctness}. We finally use the filtered dataset to finetune strong model, expecting to solve the problem of overfitting on wrong labels.
% This approach ensures the training process is grounded on accurate and reliable labels, reducing the impact of noise and improving the overall effectiveness of the weak-to-strong generalization framework.

We further analyzed the effectiveness of chain-of-thought prompting, detailed in Appendix \ref{app:cotda}.

\subsection{Stage II: Mitigating Question Degeneration}
% \subsection{Stage II: Data Refinement}
% After Stage I, we newly get a finetuned model \( M_{\text{finretune}} \) and two filtered datasets \( D_{\text{filtered}} \) and \( D_{\text{discarded}} \). For samples discarded in the first stage, their questions might suggest more difficulty or less common topics, making the weak model uncertain about the answer. However, these questions are still vital to the overall performance, as test set questions cover all ranges of difficulty and diversity. As finetuned model \( M_{\text{finretune}} \) surpasses its weak teacher in performance, we use finetuned model to refine answers for the discarded samples by generating multiple answer sets for each question. Similar to the first stage, we apply a certainty threshold filter to ensure data correctness, append filtered samples to the training set and further train the initial strong model with this updated dataset.

%SC 折线图 gsm8k+Math sc threshold&acc
Following Stage I, the finetuned model \( M_{\text{finretune}} \) and two distinct datasets are produced: a filtered dataset \( D_{\text{filtered}} \) containing high-certainty questions and a discarded dataset \( D_{\text{discarded}} \) comprising low-certainty questions. The discarded questions often represent questions with higher difficulty or less common topics, where the weak model struggled to provide confident answers. Despite this, these questions remain crucial for improving overall model performance, as the test set typically encompasses a diverse range of difficulty levels and topics.
Meanwhile, the finetuned model in Stage I, having its ability elicited by labels from weak teacher, now outperforms its weak teacher, showing the potential to solve questions beyond weak model's ability.

To address this, the finetuned student model—now exceeding the weak model in performance—is employed to generate answers for the discarded questions. For each question in the discarded question set, the finetuned model generates a variety of potential answers, providing a more accurate and comprehensive set of responses than its teacher. Similar to Stage I, an uncertainty-based filtering process is applied to retain only high-confidence samples, producing a high quality dataset, shown as "Training set B" in Figure \ref{fig:framework}.

The refined, high-certainty samples are then appended to the training set, creating an enriched dataset. This updated training set is subsequently used to finetune the initial strong model, enhancing its ability to generalize across the full spectrum of question difficulty and diversity. This refinement process ensures the inclusion of valuable but initially uncertain data, maximizing the strong model's potential and overall performance.

\section{Experiments}
\subsection{Experimental Settings}
\begin{figure*}[t]
\centering
  \includegraphics[width=\textwidth]{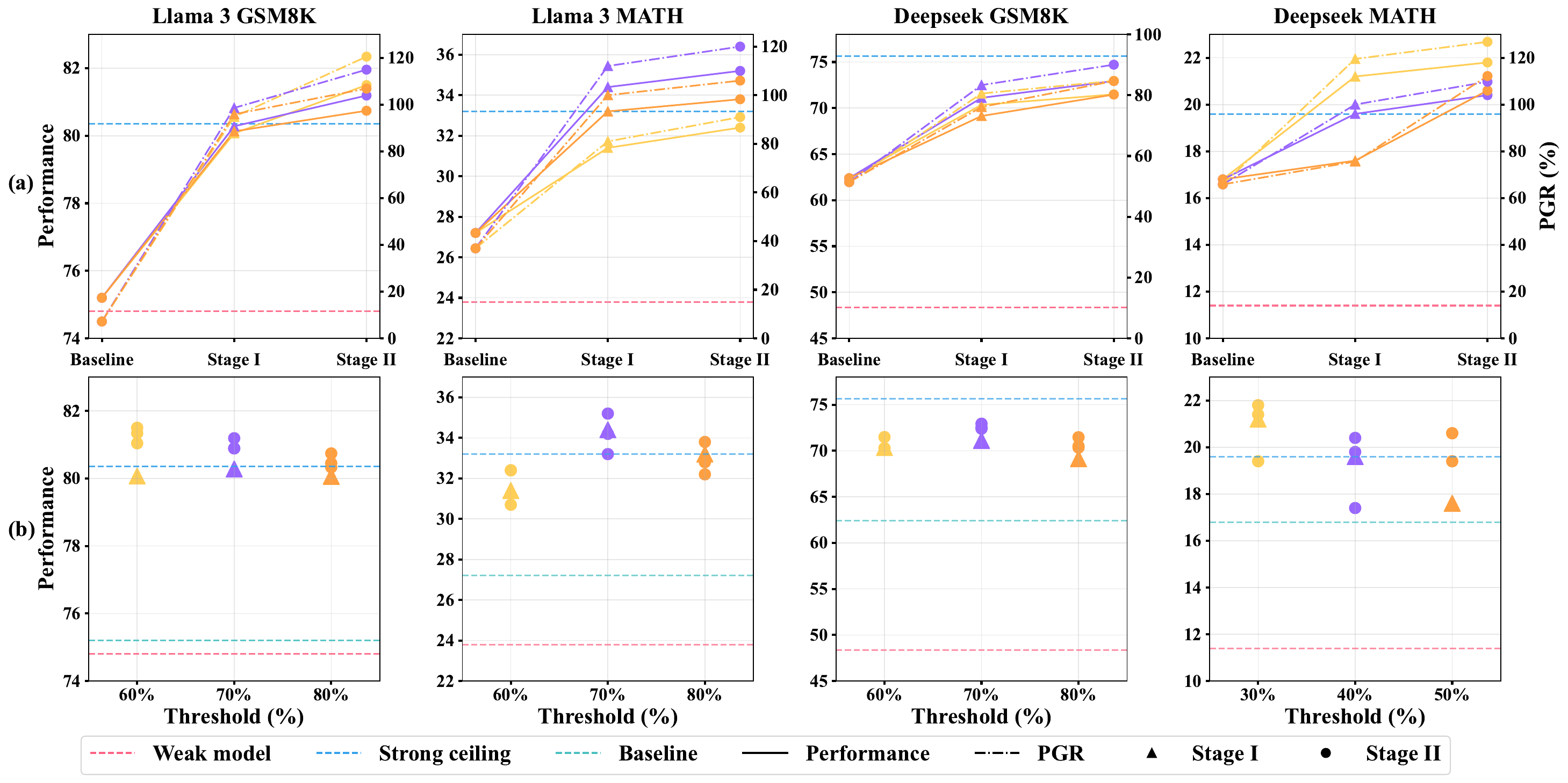}
  \caption{
  % Main experimental results across different model series (Llama 3 and Deepseek) on GSM8K and MATH datasets. 
  \textbf{(a)} The upper row shows the performance trajectory and PGR across different stages (Baseline, Stage I, and Stage II). 
  The solid lines represent model performance (left y-axis), while the dash-dotted lines show PGR values (right y-axis). 
  % Red and blue dashed lines indicate the weak model performance and strong model ceiling respectively.
  \textbf{(b)} The lower row demonstrates the impact of different filtering thresholds on model performance, with triangles representing Stage I results and circles representing Stage II results.
  For each experimental setting, points with the same color correspond to the same Stage I filtering threshold.
  Results show consistent improvement patterns across all model configurations, with Stage II generally achieving better performance than Stage I. 
  % Detailed numerical results are provided in Appendix B.
  }
  \label{fig:main_results}
  \vskip -0.2in
\end{figure*}

\paragraph{Dataset} We conduct experiments on two prominent mathematical reasoning benchmarks, the grade-school level reasoning task GSM8K \cite{gsm8k} and the more challenging MATH task \cite{hendrycksmath2021}. For training, we use the same training set as \citet{weaktostrongreasoning} for both weak model labelling and strong model training. 
% We employed chain-of-thought prompting to enhance performance for data labelling, formatted within the model’s chat template.
For evaluation, we utilized the GSM8K evaluation set, which contains 1,319 data points. For MATH, we used the smaller subset as the primary evaluation test set following \citet{math500}, which contains 500 data points.
We compared the model's performance on the 500 samples subset with that on the original test dataset, with details provided in Appendix \ref{app:Additional Analysis}.
% \cqy{Add reasons why using one-shot for llama but zero-shot for the others.}

\paragraph{Models} We use several models to investigate the effectiveness of our framework, including the Llama 3 series \cite{llama3} (Llama 3 8B Instruct, Llama 3 70B)
% the Gemma-2 series \cite{gemma2} (Gemma2-9B-Instruct, Gemma2-27B-Base),
and the Deepseek series \cite{deepseek} (Deepseek 7B Chat, Deepseek 67B Base).
% This diversity allowed us to examine the impact of our method within the same model family.

\paragraph{Evaluation Metrics} We use accuracy and performance gap recovered (PGR) as our primary evaluation metrics. For PGR, we define the performance of small instruct/chat models as "weak performance", and the performance of strong models after finetuned with golden labels as "strong ceiling", each representing the starting and the goal performance we aim to achieve.
% Both metrics were employed to assess the model's overall performance and to measure the effectiveness of the weak-to-strong generalization approach, highlighting the elicited abilities of the model and the extent to which the performance gap was recovered.
Both metrics were employed to assess the effectiveness of the weak-to-strong generalization approach, highlighting the elicited abilities of the model and the extent to which the performance gap was recovered.

\subsection{Main results}

As illustrated in Figure \ref{fig:main_results}, % our framework effectively enhances the performance of strong models and significantly narrows the performance gap between the weak model and the strong ceiling. 
our framework significantly narrows the performance gap between finetuned strong model and strong ceiling, meanwhile effectively eliciting strong model's ability.
Our experimental results demonstrate the efficacy of our framework across multiple model series, including Llama 3 and Deepseek.
% including Llama 3, Gemma-2, and Deepseek. 
For the Llama 3 model, specifically the 70B variant, the performance in weak-to-strong generalization (PGR) on the GSM8K dataset shows a remarkable improvement, rising from 7.19\% to 120.50\% when utilizing the smaller Llama 3 8B Instruct model as the weak model. This improvement is accompanied by an increase in task performance, which climbs from 75.20\% to 81.50\%. Similar enhancements are observed on the MATH dataset, where PGR increases from 36.17\% to 121.28\% and task performance rises from 18.2\% to 35.2\%.

Comparable gains are seen with the Deepseek model series. On the GSM8K dataset, PGR increases significantly from 51.39\% to 90.04\%, while task performance improves from 62.39\% to 72.94\%. For the MATH dataset, PGR improves from 65.85\% to 126.83\%, with performance rising from 16.8\% to 21.8\%.
% For Gemma-2 on GSM8K, performance improves from 67.62\% to 72.32\% (PGR from 2.06\% to 66.62\%), and for Deepseek on GSM8K, performance increases from 62.39\% to 72.94\% (PGR from 51.39\% to 90.04\%). On the MATH dataset, Deepseek’s performance improves from 16.8\% to 21.8\% (PGR from 65.85\% to 126.83\%).

% Stage I Results

\subsection{Performance Gains from Enhanced Supervision Quality}%Results of Stage I}
% \cqy{Rename}
As illustrated in Figure \ref{fig:main_results}(a), the uncertainty-based filtering approach implemented in Stage I consistently outperforms the conventional baseline across multiple datasets and model configurations. Specifically, for Llama 3 on the GSM8K dataset, the weak-to-strong generalization performance improves substantially from 7.19\% to 98.56\% in PGR, accompanied by an increase in task performance from 75.20\% to 80.28\%. On the MATH dataset, PGR rises from 36.17\% to 112.77\%, while task performance increases from 18.2\% to 34.0\%. 
Similarly, for Deepseek on GSM8K, PGR increases from 51.39\% to 83.33\%, while performance enhances from 62.39\% to 71.11\%. On the MATH dataset, Deepseek shows a notable improvement, with PGR rising from 65.85\% to 119.51\%, and task performance increasing from 16.8\% to 21.2\%.

% Stage II Results
\subsection{Further Improvement from Enhanced Question Quality}
% \subsection{Results of Stage II}
% \cqy{Rename}
As further illustrated in Figure \ref{fig:main_results}(b), the refinement process in Stage II effectively enhances the quality of the training data, particularly in terms of difficulty and diversity, leading to significant improvements in model performance. Specifically, for the Llama 3 series, the strong model achieves a peak PGR of 120.50\% on the GSM8K dataset, reflecting an additional 21.94\% improvement compared to the finetuned strong model in Stage I, corresponding to a performance of 81.50\%. On the MATH dataset, we observe a peak PGR of 121.28\%, with a further increase of 8.51\% compared to Stage I, reaching 35.2\% on task performance.

For the Deepseek series, the strong model attains a peak PGR of 90.04\% on GSM8K, marking an additional 6.71\% improvement over Stage I, with a corresponding finetuned performance of 72.94\%. On MATH, the peak PGR reaches 126.83\%, demonstrating a further increase of 7.32\% compared to Stage I, with task performance reaching 21.8\%.

% \subsection{Conclusion}
% These results collectively highlight the effectiveness of our framework in significantly enhancing weak-to-strong generalization. By incorporating a two-stage process—first purify supervision by filtering high-uncertainty labels and then further mitigate degeneration by refining the data's difficulty and diversity—our approach improves both the supervision quality and question quality of the training labels, ultimately leading to more robust model performance.

% As also shown in Figure X(c), the data refinement process in Stage II effectively bootstraps both the difficulty and diversity of the data, resulting in enhanced model performance. In the case of Llama 3, the strong model achieves a peak performance of 81.50\% (PGR of 50\%) on GSM8K and 35.2\% (PGR of 60\%) on MATH. For Gemma-2 on GSM8K, the strong model reaches a maximum finetuned performance of 72.32\% (PGR of 40\%), and for Deepseek on GSM8K, the best finetuned performance is 72.94\% (PGR of 40\%), with a performance of 21.8\% (PGR of 60\%) on MATH.

% These results collectively demonstrate that weak-to-strong generalization can be significantly improved by our framework, both by filtering out high-uncertainty labels and further bootstrapping quality, including difficulty and diversity, thereby increasing the correctness and reliability of the labels.
 
\section{Analysis}
\subsection{The Impact of Excessive Filtering on Supervision Quality}

As shown in Figure \ref{fig:sc_correctness}, label correctness increases as model uncertainty decreases. However, in preliminary experiments during Stage I, we observed an intriguing trend: while performance improves initially as uncertainty decreases, it starts to deteriorate after a certain threshold. This suggests that other factors, beyond supervision quality, influence weak-to-strong generalization, and existing filtering methods may have inherent limitations.

\begin{figure}[t]
\centering
  \includegraphics[width=\columnwidth]{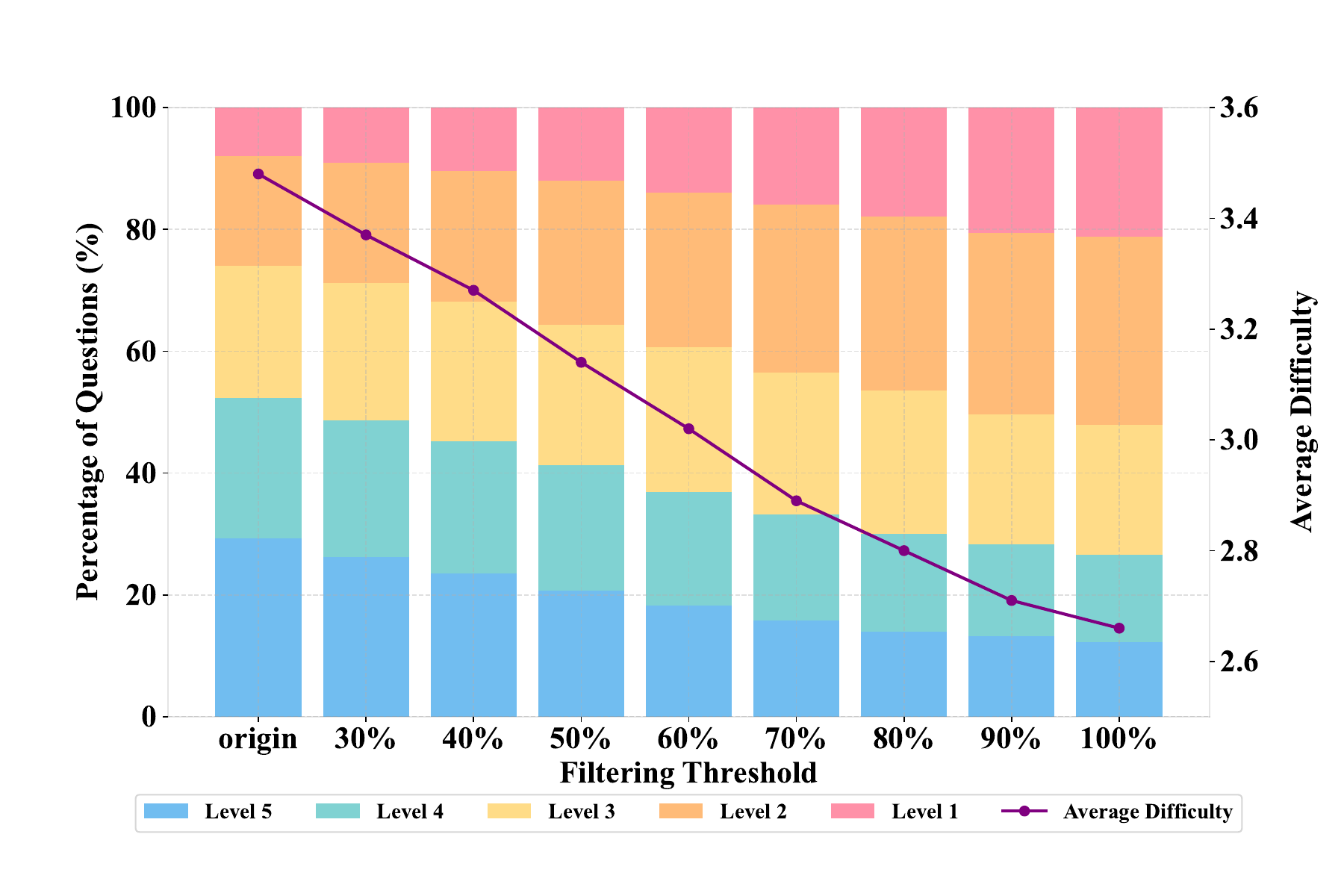}
  \caption{Impact of filtering threshold on question difficulty distribution. As the threshold increases, the proportion of difficult questions (Levels 4-5) decreases, while easier questions (Levels 1-2) increase, resulting in a decline in average difficulty from 3.48 to 2.66.}
  \label{fig:diff1}
\end{figure}

\textbf{Reduction in Data Difficulty}  
Figure \ref{fig:diff1} shows that increasing the filtering threshold leads to a decrease in average difficulty, with fewer hard questions (Levels 4-5) remaining in the dataset. These harder questions represent areas where the weak model is less confident, suggesting they are beyond its current capabilities. In contrast, easier questions (Levels 1-2), where the model is more confident, dominate the dataset. This results in a less challenging training set, hindering the model's ability to generalize to more difficult problems and contributing to data degeneration.

\begin{figure}[t]
\centering
  \includegraphics[width=\columnwidth]{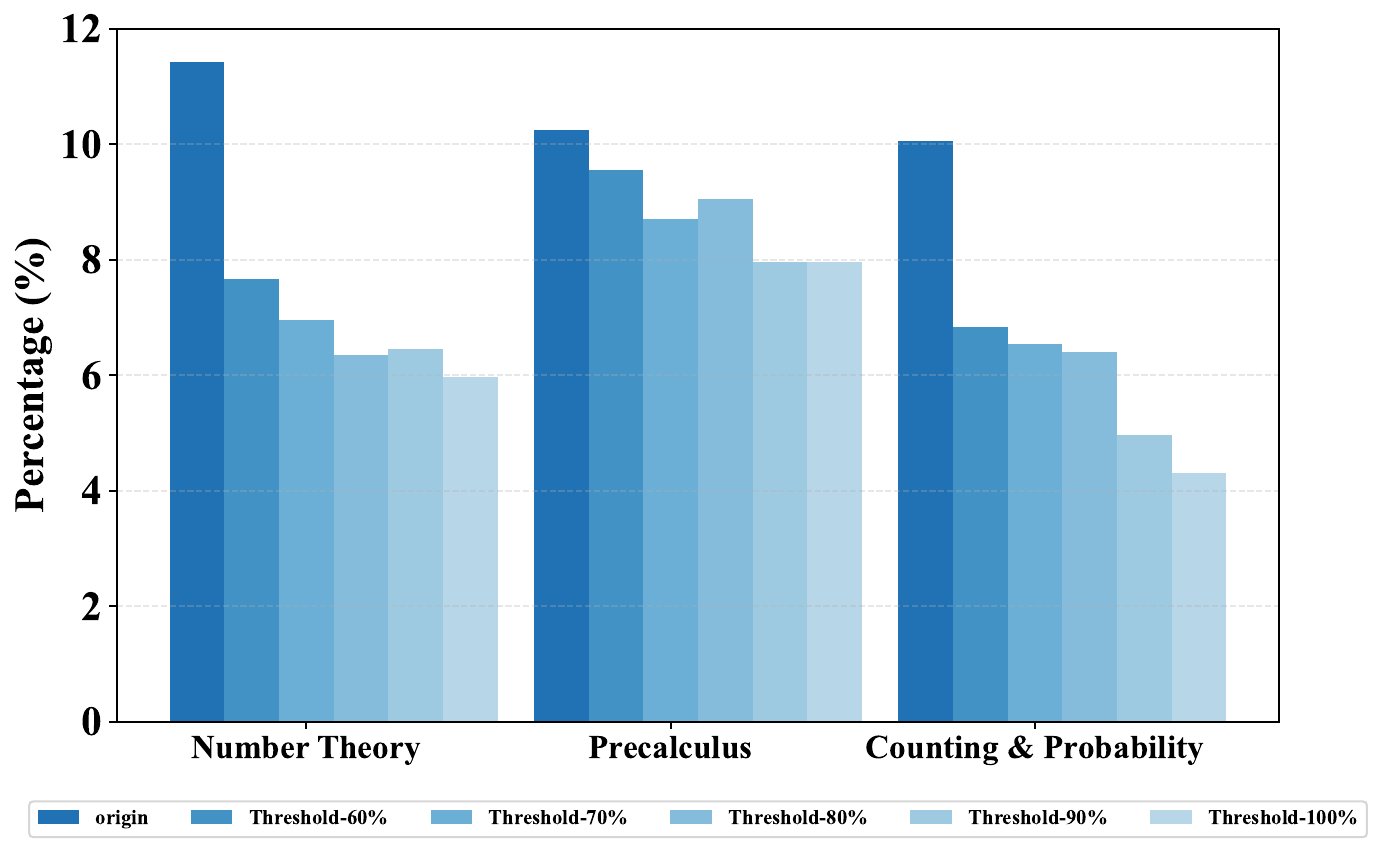}
  \caption{Changes in topic distribution across filtering thresholds for three representative mathematical categories. Filtering causes shifts in topic distribution, with minor categories seeing more reductions.
  % A detailed analysis of all categories is provided in Appendix \ref{appendix}.
  }
  \label{fig:div1}
  \skip -0.2in
\end{figure}

\textbf{Shift in Data Diversity}  
As shown in Figure~\ref{fig:div1}, filtering also causes a significant shift in the diversity of questions. For instance, the Counting and Probability section drops from 10.79\% to 4.31\%, reflecting changes in the model's uncertainty. This shift in data diversity impacts the variety of question types, reducing exposure to harder topics. The complete trends and numerical results across all categories are provided in Appendix \ref{app:div_whole}.
% \cqy{empty ref}

Once the filtering threshold surpasses a certain point, performance degrades due to the exclusion of important, challenging data. While reducing label uncertainty can improve performance, excessive filtering diminishes the dataset's diversity, particularly regarding difficulty and topic variety. This limits the model's ability to generalize effectively, leading to degeneration in its overall performance.

\subsection{The Robust Effectiveness of Data Refinement in Stage II}
% concise ver
To address excessive filtering, we propose a strategy that balances uncertainty-based filtering with the preservation of question quality, including difficulty and diversity. In Stage II, we regenerate answers for discarded questions from Stage I using the finetuned model, filtering them by uncertainty before adding low-uncertainty samples to the dataset.

As shown in Figure \ref{fig:main_results}(a), Stage II consistently improves performance across all filtering thresholds, demonstrating the effectiveness of our framework in recovering lost data and boosting performance.

\begin{figure}[t]
    \centering
    \begin{subfigure}{\columnwidth}
        \includegraphics[width=\columnwidth]{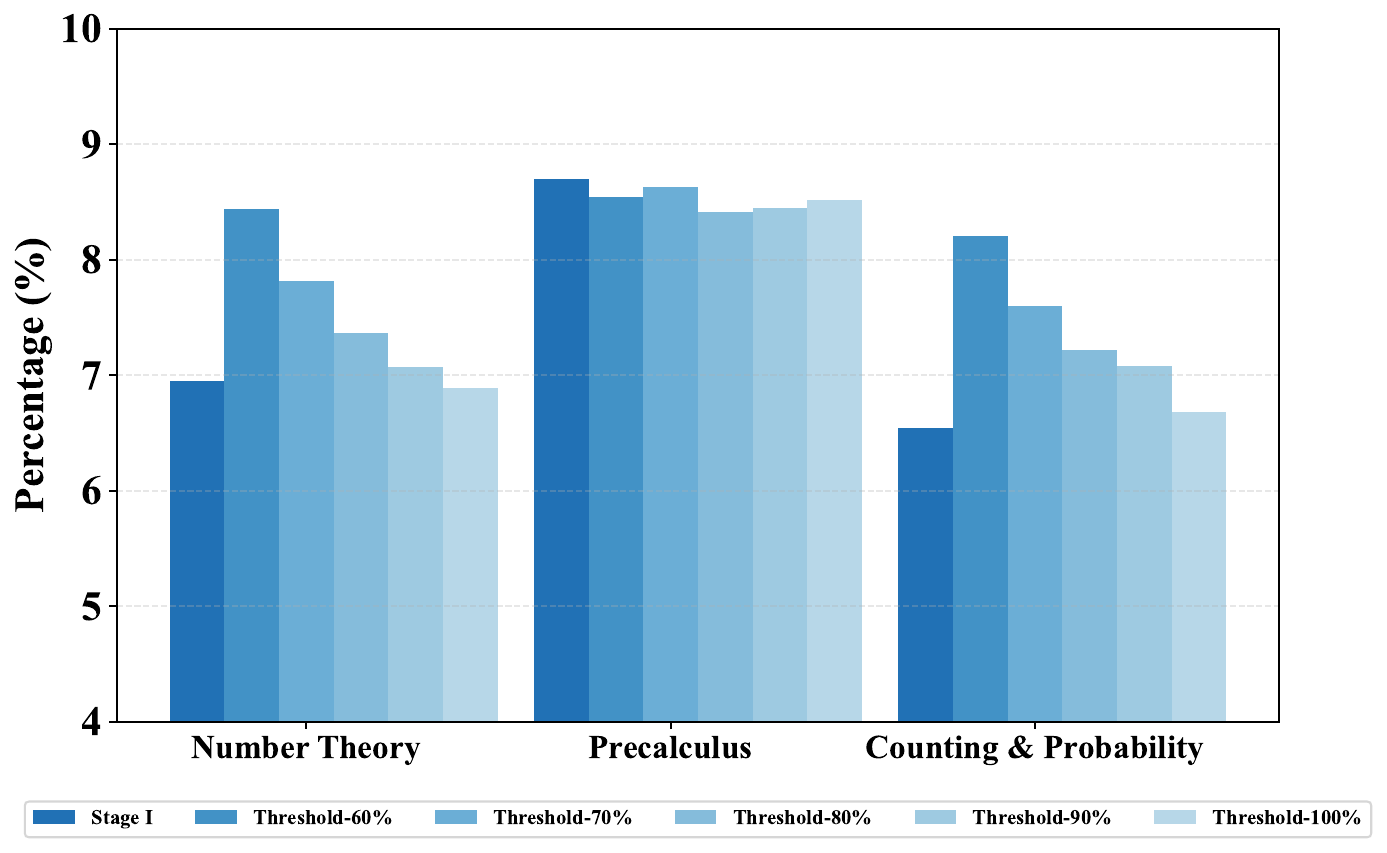}
        \caption{Topic distribution comparison in Stage II under different thresholds.}
        \label{fig:div2}
    \end{subfigure}
    \begin{subfigure}{\columnwidth}
        \includegraphics[width=\columnwidth]{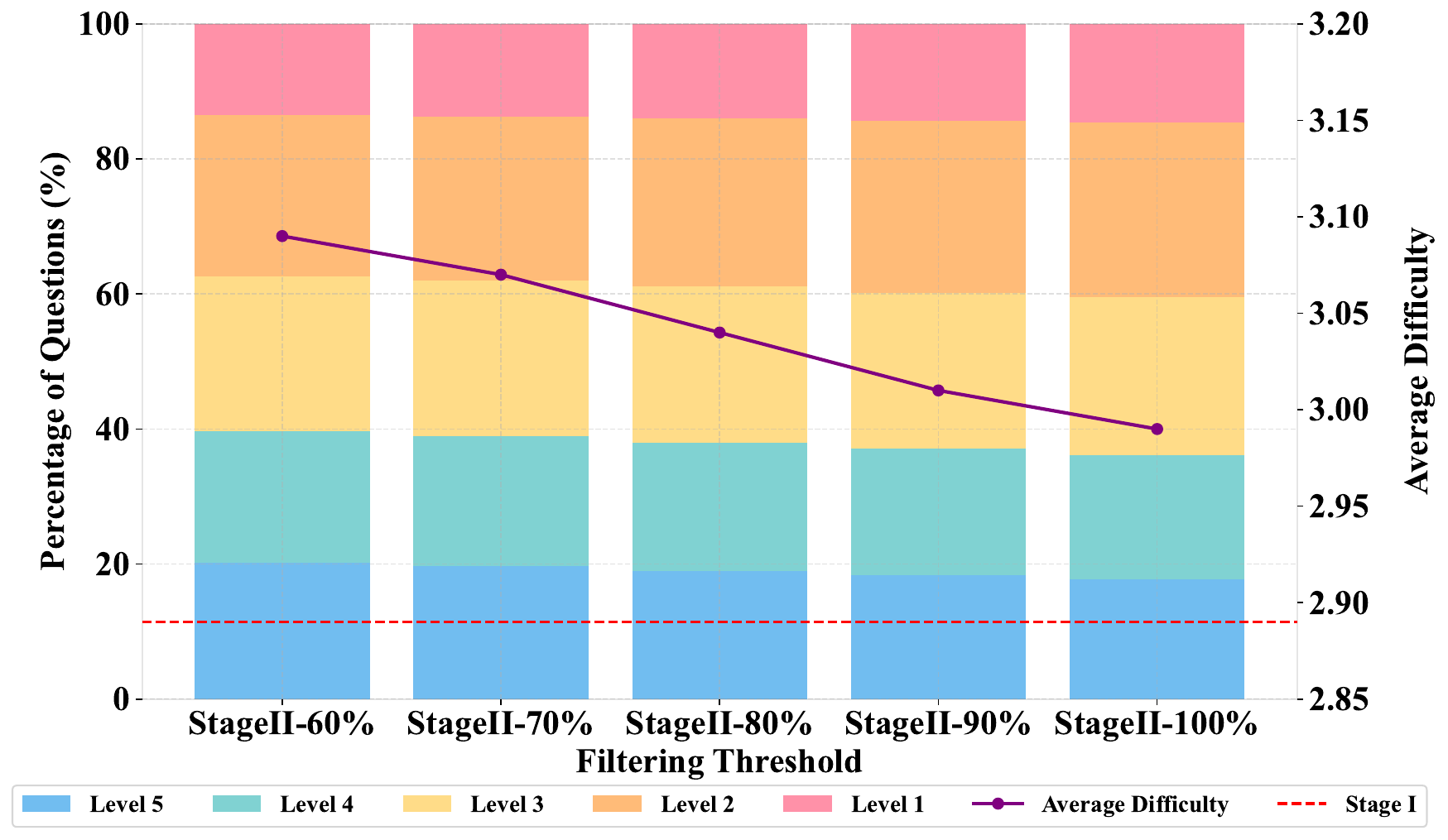} 
        \caption{Distribution of difficulty levels and average difficulty scores in Stage II.}
        \label{fig:diff2}
    \end{subfigure}
    \caption{Difficulty and diversity analysis in Stage II (GSM8K, Llama 3, Threshold-70\%), showing improved preservation of question quality.}
    \label{fig:stage2_analysis}
    \vskip -0.2in
\end{figure}

Figure \ref{fig:stage2_analysis} shows recovery in both difficulty and diversity, with the refined dataset closely resembling the original. For Llama 3 on MATH, PGR increases from 112.77\% to 121.28\%, and performance rises from 34.4\% to 35.2\%. Similar results are observed in Figure \ref{fig:main_results}, highlighting the framework’s robustness across models and datasets.

Additionally, Figure \ref{fig:main_results} demonstrates that even models with initially lower performance show significant improvements. For the Deepseek series on MATH, the performance gap between thresholds narrows in Stage II, indicating that the framework effectively recovers discarded data from over-filtered scenarios while refining fewer under-filtered questions.

\subsection{The Importance of Label Filtering in Stage II}
% concise ver
In Stage II, we focus on enhancing question quality and mitigating degeneration by using the finetuned model to generate answers for discarded questions from Stage I. Instead of adding all generated answers back, we apply an uncertainty-based filter to ensure only reliable answers are reintegrated, preventing the inclusion of low-quality data.

Table \ref{tab:filter ablation in Stage II} summarizes the results of the ablation study comparing the framework with and without the filtering process, using the Llama 3 model series on the GSM8K dataset.

\begin{table}[H]
\centering
\small
\begin{tabular}{lccc}
\hline
             & Origin                                          & With Filter & Without Filter \\ \hline
Stage I-50\% & 78.99                                           & 80.89 (+1.90) & 78.31 (-0.68)  \\
Stage I-60\% & 80.07                                           & 81.50 (+1.43) & 78.84 (-1.23)  \\
Stage I-70\% & 80.28                                           & 81.19 (+0.91) & 80.28 (+0.00)  \\
Stage I-80\% & 80.06                                           & 80.74 (+0.68) & 79.59 (-0.47)  \\ \hline
\end{tabular}
\caption{The impact of \textbf{With} vs. \textbf{Without} label filtering in Stage II on Weak-to-Strong Generalization.}
\label{tab:filter ablation in Stage II}
\vskip -0.2in
\end{table}

As shown in Table \ref{tab:filter ablation in Stage II}, appending all generated samples without filtering leads to performance degradation, highlighting that indiscriminate inclusion reduces supervision quality. The uncertainty-based filter ensures optimal supervision and question quality, which are critical for effective weak-to-strong reasoning generalization.

\subsection{Exploring the Potential for Further Iterative Refinement}
% concise ver
While our current framework demonstrates considerable effectiveness, we recognize that additional iterations could further improve question quality, thereby enhancing overall framework performance. Specifically, the refinement process in Stage II—where discarded questions are recovered and answered using the finetuned strong model—holds significant potential for further improvement. This iterative process, as the model’s ability improves, may offer a pathway for continuous enhancement of question quality.

\begin{table}[t]
\centering
\resizebox{\columnwidth}{!}{
\begin{tabular}{lcc}
\hline
                          & Accuracy    & PGR                      \\ \hline
\multicolumn{3}{l}{GSM8K}                                          \\ \hline
Baseline                  & 62.39       & 51.39\%                  \\
Stage I                   & 71.11       & 83.33\% (+31.94\%)       \\
Stage II                  & 72.94       & 90.04\% (+38.65\%)       \\
Stage Exp-Threshold-80\%  & 72.26       & 87.55\%                  \\
Stage Exp-Threshold-90\%  & 72.93       & 90.00\%                  \\
Stage Exp-Threshold-100\% & \textbf{{\ul 73.77}} & \textbf{{\ul 93.08\% (+41.69\%)}} \\ \hline
\multicolumn{3}{l}{MATH}                                           \\ \hline
Baseline                  & 16.8        & 65.85\%                  \\
Stage I                   & 21.2        & 119.51\% (+53.66\%)      \\
Stage II                  & 21.8        & 126.83\% (+60.98\%)      \\
Stage Exp-Threshold-50\%  & 21.4        & 120.71\%                 \\
Stage Exp-Threshold-40\%  & 21.2        & 119.51\%                 \\
Stage Exp-Threshold-30\%  & \textbf{{\ul 22.4}}  & \textbf{{\ul 134.15\% (+68.3\%)}} \\ \hline
\end{tabular}
}
\caption{Performance comparison of iterative refinement on GSM8K and MATH datasets (Deepseek model). Best results are underlined.}
\label{tab:stage Exp}
\vskip -0.2in
\end{table}

We introduce an additional iteration, which we term Stage Exp, aimed at refining discarded questions by utilizing finetuned strong model in Stage II to generate answers, and append samples to the existing dataset after uncertainty filtering. Due to computational limits, Stage Exp experiments focused on Deepseek series with best configurations for GSM8K and MATH.

% Due to computational constraints, Stage Exp experiments were conducted on Deepseek series, focusing on best-performing configurations for GSM8K and MATH datasets. 

As shown in Table \ref{tab:stage Exp}, our framework demonstrates a promising potential for further refinement by leveraging the power of finetuned strong models to iteratively enhance discarded questions. However, it is important to acknowledge that the selection of an optimal threshold for these further iterations remains an open question, which we intend to address in future work.

\section{Related Work}

% \subsection{LLM-driven Synthetic data}
% Through the development of LLM, data quality has been a long standing key problem, as it greatly affects the final performance. As model's performance becomes stronger, using model generated data as training data has been seen as a promising approach. 

\subsection{AI Deceptions}
% noisy labels might be a reason
% concise ver
A persistent challenge in weak-to-strong generalization is AI deception, where strong models overfit to noisy labels from weak models, hindering their ability to generalize to complex samples~\cite{Superficialalignment}. A similar issue in reinforcement learning from human feedback (RLHF) is identified by \citet{rlhfmisleadhuman}, where models mislead human evaluators. 
To address this, they propose the "U-SOPHISTRY" pipeline.

This behaviour is akin to model sycophancy, where models align with human feedback at the expense of accuracy. Early work by \citet{sycophancycotra2021} and \citet{sycophancymodelwrittenevaluations} shows models often aim to please users. \citet{understandingsycophancy} attributes this to human preference biases. Solutions such as synthetic data~\citep{syntheticreducesycophancy} and pinpoint tuning~\citep{sycophancypinpointtuing} aim to mitigate sycophancy, while \citet{sycophancyuncertainty} links it to model uncertainty.

\subsection{Weak-to-Strong Generalization}
% concise ver
Weak-to-strong generalization, introduced by OpenAI \cite{OAIweaktostrong}, has led to advancements in model training and supervision. Recent studies explore ensemble learning to improve labels by integrating predictions from smaller models \cite{cosupervisedlearningweaktostrong,ensemw2s,bayesianweakstostrong}. In terms of training methodologies, \citet{contransweaktostrong} replaces traditional sample-label pairs with concept vectors to enhance learning representations, while \citet{reliabilityawareweaktostrong} introduces filtering mechanisms and confidence-based reweighting strategies. Furthermore, a two-stage learning framework presented in \citet{weaktostrongreasoning} iteratively refines training data, \citet{w2ssearch} enhances strong model with weak test-time guidance, and \citet{macpoweaktostrong} proposes a multi-agent contrastive preference optimization approach. Theoretical foundations of weak-to-strong generalization have been studied\cite{theoreticalanalysisweaktostrong,quantifyinggainweaktostrong,provableweaktostrong}.
% In addition to methodological advancements, several studies investigate the theoretical foundations of weak-to-strong generalization\cite{theoreticalanalysisweaktostrong,quantifyinggainweaktostrong,provableweaktostrong}. 
Safety considerations are also highlighted, addressing AI safety implications within weak-to-strong frameworks \cite{Superficialalignment,weaktostrongbackdoorattack,weaktostrongsafetypilot}.

\section{Conclusion}
In this paper, we introduce a two-stage training framework to enhance weak-to-strong generalization through mitigating overfitting. By focusing on both supervision and question quality, we demonstrate that traditional data filtering methods, while improving supervision, can reduce question difficulty and diversity. 
Our framework mitigates this by relabeling discarded questions using the finetuned strong model, maintaining both supervision accuracy and question quality.

Experiments on the GSM8k and MATH benchmarks demonstrate that our approach significantly outperforms conventional weak-to-strong generalization methods, improving the performance gap recovered (PGR). This validates the effectiveness of our framework in addressing overfitting and enhancing model capabilities on challenging tasks.

% Future work will explore further improvements in filtering techniques and extend the framework to other domains including alignment and more general senarios.

\section*{Limitations}
Our experiments demonstrate strong performance on mathematical reasoning tasks, though the framework's effectiveness remains to be validated across other domains. Through extensive experimentation, we identified optimal confidence thresholds for filtering model predictions. However, these thresholds vary significantly across different tasks and datasets, making automatic threshold selection an important direction for future research. Additionally, the computational overhead of our two-stage finetuning approach, particularly in the second stage, may pose scalability challenges for large-scale applications or real-time scenarios.

\bibliography{custom}

\newpage

\appendix
% \section{Experiment Details}
% \section{Theroretical Proof}
% \subsection{Theoretical Analysis of Weak-to-Strong Generalization}
% According to the expansion theory proposed by \citet{theoreticalanalysisweaktostrong}, weak-to-strong generalization emerges due to two important effects: pseudolabel correction and coverage expansion. 

\section{Dataset details}
\label{apx:dataset_details}
\subsection{Dataset Statistics}
For the original question set used in GSM8K and MATH, we followed the methodology of \citet{weaktostrongreasoning}, adopting the same training set for both datasets. Specifically, we used their dataset $D_2$, which was employed for training the Llama 2 70B model. For GSM8K, the dataset consists of 7,000 samples, while for MATH, the dataset comprises 6,000 samples.

For evaluation, we utilized the original evaluation set for GSM8K and the test set from \citet{math500}, which contains 500 samples. We compared the model's performance on the 500 samples subset with that on the original test dataset, with details provided in Appendix \ref{app:Additional Analysis}.

\subsection{Implementation Details} 
For answer generation within the framework, we utilize chain-of-thought (CoT) prompting, as its necessity has been outlined in Section 5.4. In Stage I, answers are generated using zero-shot CoT prompting for the weak models in the Deepseek series. However, for the Llama 3 series, we observed that the Llama 3 8B Instruct model performed below expectations, prompting us to switch from zero-shot to one-shot CoT to enhance its performance.

For sampling parameters, we generate answers with a temperature of 0.6 and top-p of 0.9 for uncertainty-based filtering to ensure diverse and coherent outputs, while using greedy decoding during evaluation to enhance stability.

In both Stage II and the experimental Stage Exp, discussed in Section 5.5, all answers are generated using zero-shot prompting. During the filtering process, after excluding answers based on model confidence, we also discard responses that fail to generate valid answers or do not adhere to the CoT format.

% The impact of these two filtering aspects is further discussed in Appendix x.x.
% \subsection{Prompting Templates}

% Below are several prompting templates used for generating responses, each tailored to different models or requirements:

% \subsubsection{Direct Template}
% This template is straightforward and minimalistic.
% \begin{verbatim}
% Question: [INPUT]
% Answer:
% \end{verbatim}

% \subsubsection{Llama3-GSM8K Template}
\subsection{Prompting Template}
To better evaluate and compare the mathematical reasoning capabilities of different models, we designed specific prompting templates. For Stage I answer generation, we employ chat-style templates to facilitate more natural responses, while in Stage II answer generation and evaluation, we utilize the direct template for standardization.

We designed the following prompting templates for different models, where [INPUT] denotes the mathematical question to be solved.

\noindent\textbf{Direct Template:}
% \begin{small}
% \begin{verbatim}
% Question: [INPUT]
% Answer:
% \end{verbatim}
% \end{small}

\begin{figure}[!ht] 
\begin{AIbox}{Direct Template:}
{\color{blue}\bf Prompt:} \\
Question: [INPUT] \\
Answer:
\end{AIbox} 
\end{figure}

\noindent\textbf{Llama 3 GSM8K Template:}

\begin{figure}[!ht] 
\begin{AIbox}{Llama 3 GSM8K Template:}
{\color{blue}\bf Prompt:} \\
<|begin\_of\_text|> \\
<|start\_header\_id|>user<|end\_header\_id|> \\
Please additionally write your final answer with \#\#\#\#, like the example: \\
Question: Greg has his own dog walking business. He charges \$20 per dog plus \$1 per minute per dog for walking the dog. If he walks one dog for 10 minutes, two dogs for 7 minutes and three dogs for 9 minutes, how much money, in dollars, does he earn? \\
Answer: Greg earns \$20 + \$1 x 10 minutes = \$21 for walking the first dog. He earns \$20 + \$1 x 7 minutes = \$27 for walking the second dog. He earns \$20 + \$1 x 9 minutes = \$29 for walking the third dog. Therefore, Greg earns \$21 + \$27 + \$29 = \$77 for walking the three dogs. \#\#\#\# 77 \\ 
Question:  \\
Answer: \\
<|eot\_id|> \\
<|start\_header\_id|>assistant<|end\_header\_id|> 
\end{AIbox} 
\end{figure}

% \begin{small}
% \begin{verbatim}
% <|begin_of_text|>
% <|start_header_id|>user<|end_header_id|>
% Please additionally write your final answer with ####, 
% like the example:
% Question: Greg has his own dog walking business. 
% He charges $20 per dog plus $1 per minute per dog 
% for walking the dog. If he walks one dog for 10 
% minutes, two dogs for 7 minutes and three dogs for 
% 9 minutes, how much money, in dollars, does he earn?
% Answer: Greg earns $20 + $1 x 10 minutes = $21 
% for walking the first dog.
% He earns $20 + $1 x 7 minutes = $27 for walking 
% the second dog.
% He earns $20 + $1 x 9 minutes = $29 for walking 
% the third dog.
% Therefore, Greg earns $21 + $27 + $29 = $77 for 
% walking the three dogs.
% #### 77
% Question: 
% Answer:
% <|eot_id|>
% <|start_header_id|>assistant<|end_header_id|>
% \end{verbatim}
% \end{small}

\noindent\textbf{Llama 3 MATH Template:}

\begin{figure}[!ht] 
\begin{AIbox}{Llama 3 MATH Template:}
{\color{blue}\bf Prompt:} \\
<|begin\_of\_text|> \\
<|start\_header\_id|>user<|end\_header\_id|> \\
Answer the math question step by step. Our answers need to end with 'The answer is '. \\
Question: [INPUT] \\
Answer: Let's think step by step. \\
<|eot\_id|> \\
<|start\_header\_id|>assistant<|end\_header\_id|>
\end{AIbox} 
\end{figure}

% \begin{small}
% \begin{verbatim}
% <|begin_of_text|>
% <|start_header_id|>user<|end_header_id|>
% Answer the math question step by step. 
% Our answers need to end with 'The answer is '.
% Question: [INPUT]
% Answer: Let's think step by step.
% <|eot_id|>
% <|start_header_id|>assistant<|end_header_id|>
% \end{verbatim}
% \end{small}

\newpage
\noindent\textbf{DeepSeek Templates:}

\begin{figure}[!ht] 
\begin{AIbox}{DeepSeek Templates:}
{\color{blue}\bf Prompt:} \\
<|begin\_of\_sentence|> \\
User: Question: [INPUT] \\
Please reason step by step, and put your final answer after 'The answer is: '. \\
Assistant:
\end{AIbox} 
\end{figure}

% \begin{small}
% \begin{verbatim}
% <|begin_of_sentence|>
% User: Question: [INPUT]
% Please reason step by step, and put your final 
% answer after 'The answer is: '.
% Assistant:
% \end{verbatim}
% \end{small}

\section{Training Details}
For the supervised finetuning in our framework, we perform full-parameter finetuning on the strong model. The finetuning is carried out with a learning rate of $1 10^{-5}$, a warmup ratio of 0.1, and a cosine learning rate scheduler. We use a batch size of 128 and train for 2 epochs on both the GSM8K and MATH datasets. The implementation is based on the LlamaFactory~\citep{llamafactory} framework and all experiments are conducted using 64 H100 80GB GPUs to ensure efficient processing and model optimization.

\section{Additional Analysis}
\subsection{Theoretical Analysis}
The weak-to-strong generalization phenomenon can be theoretically explained through two key mechanisms: pseudolabel correction and coverage expansion, as demonstrated by \citet{theoreticalanalysisweaktostrong}. These mechanisms enable the student model to both correct erroneous labels from the weak teacher and generalize to samples where the teacher lacks confidence.

Let us consider a student model $f$, a covered subset $S$ partitioned into correct samples $S^{good}$ and incorrect samples $S^{bad}$, and the weak teacher's error rate $\alpha$. We can establish the relationship between the gold error $\err(f,y|S)$ and the weak error $\err(f,\tilde{y}|S)$ of the student model.

Given that $\mathcal{M}'(S^{good}, \mathcal{F})$ satisfies $(c,q)$-expansion on $(S^{bad}, S^{good})$ where $q < \frac{3}{4}(1-2\alpha)$, and for an optimal classifier $f$ whose probability of prediction errors or non-robustness is bounded by $(1-\alpha+3c\alpha)/4$, we can establish the following bound:

% \begin{equation*}
% \resizebox{.98\hsize}{!}{
% \begin{split}
% \err(f,y|S) \leq \frac{2\alpha}{1-2\alpha}\mathbb{P}(\overline{R}(f)|S) + \err(f,\tilde{y}|S) + \alpha\left(1-\frac{3}{2}c\right).
% \end{split}
% }
% \end{equation*}

\begin{equation*}
\resizebox{.98\hsize}{!}{$
\begin{split}
\err(f,y|S) \leq \frac{2\alpha}{1-2\alpha}\mathbb{P}(\overline{R}(f)|S) + \err(f,\tilde{y}|S) + \alpha\left(1-\frac{3}{2}c\right).
\end{split}
$}
\end{equation*}

This bound demonstrates that when the expansion coefficient $c$ is sufficiently large and both $\err(f,\tilde{y}|S)$ and $\mathbb{P}(\overline{R}(f)|S)$ are minimal, the true error $\err(f,y|S)$ can be significantly lower than the weak teacher's error rate $\alpha$, indicating successful pseudolabel correction.

For the uncovered set $T$, assuming $\mathcal{M}(T, \mathcal{F})$ satisfies $(c,q)$-expansion on $(S^{good}, T)$ and $\mathcal{M}'(T, \mathcal{F})$ satisfies $(c,q)$-expansion on $(S^{bad}, T)$, we can derive another bound. For a classifier $f \in \mathcal{F}$ that demonstrates good fit to weak labels on $S$ and maintains robustness on $T$ such that $\err(f,\tilde{y}|S) + \mathbb{P}(\overline{R}(f)|T) < c(1-q-\alpha)$, we have:

\begin{equation*}
\resizebox{.98\hsize}{!}{$
\begin{split}
\err(f,y|T) \leq \left(1 + \frac{\alpha}{1-2\alpha}\right)\mathbb{P}(\overline{R}(f)|T) + \max\left(q, \frac{\err(g,\tilde{y}|S) - c\alpha}{c(1-2\alpha)}\right).
\end{split}
$}
\end{equation*}

This bound becomes particularly tight when $f$ exhibits strong performance on weak labels in $S$ and maintains robustness across $T$, resulting in small values for both $\err(f,\tilde{y}|S)$ and $\mathbb{P}(\overline{R}(f)|T)$. Combined with the previous bound on $\err(f,y|S)$, these results theoretically justify the student model's capacity to surpass its weak teacher.

In our framework, the filtering mechanism serves to reduce the weak teacher's error rate $\alpha$ by excluding low-confidence samples, thereby improving supervision quality and consequently reducing $\err(f,y|S)$. However, this improvement relies on the assumption that sets $S$ and $T$ maintain similar distributional characteristics for effective generalization. As the filtering threshold increases, the shrinking of set $S$ and expansion of set $T$ can lead to distributional shifts that violate this assumption. To address this challenge, Stage II of our framework employs the finetuned model to generate predictions for previously discarded questions, selectively reincorporating high-confidence predictions into the dataset. This approach effectively reduces the weak error rate $\alpha$ while preserving distributional similarity, ultimately enhancing weak-to-strong generalization.

\subsection{The Role of Chain-of-Thought in Weak-to-Strong Reasoning}
\label{app:cotda}

In contrast to the original weak-to-strong generalization framework proposed by \cite{OAIweaktostrong}, where all tasks are classification-based, reasoning tasks like GSM8K and MATH consist of open-ended questions that lack definitive answer sets. Previous work has utilized chain-of-thought prompting to enhance performance \cite{reliabilityawareweaktostrong, weaktostrongreasoning}. This raises the question: \textbf{Can weak-to-strong generalization remain effective without chain-of-thought prompting?}

To explore this, we replicate the same baseline settings, comparing using chain-of-thought answers to manually constructed direct answers. The results are shown in Table \ref{tab:cot-da-ablation}.

% \begin{table}[]
% \begin{tabularx}[width=\columnwidth]{lcccc}
% \hline
%                        & Weak Model & Strong Ceiling & Weak-to-Strong & PGR                \\ \hline
% GSM8K Chain-of-Thought & 74.8       & 80.36          & 75.2           & 7.19\%             \\
% GSM8K Direct Answer    & 14.6       & 30.93          & 13.64          & -5.87\% (-13.06\%) \\ \hline
% MATH Chain-of-Thought  & 23.8       & 33.2           & 27.2           & 36.17\%            \\
% MATH Direct Answer     & 12.8       & 17.2           & 11.4           & -31.8\% (-67.97\%) \\ \hline
% \end{tabular}
% \end{table}

\begin{table}[t]
\resizebox{\columnwidth}{!}{
\begin{tabular}{lcc}
\hline
               & Chain-of-Thought & Direct Answer     \\ \hline
\multicolumn{3}{l}{GSM8K}                                          \\ \hline
Weak Model     & 74.8             & 14.6              \\
Strong Ceiling & 80.36            & 30.93             \\
Weak-to-Strong & 75.2             & 13.64             \\
PGR            & 7.19\%           & -5.87\%(-13.06\%) \\ \hline
\multicolumn{3}{l}{MATH}                                          \\ \hline
Weak Model     & 23.8             & 14.6              \\
Strong Ceiling & 33.2             & 30.93             \\
Weak-to-Strong & 27.2             & 11.4              \\
PGR            & 36.17\%          & -31.8\%(-76.97\%) \\ \hline
\end{tabular}
}
\caption{Performance comparison between chain-of-thought and direct answer approaches in weak-to-strong generalization on GSM8K and MATH datasets with Deepseek series.}
\label{tab:cot-da-ablation}

\end{table}

When omitting chain-of-thought prompting, we fail to observe generalization in strong models, as finetuned strong models perform worse than their weak teachers. This can be attributed to the fact that chain-of-thought prompting facilitates step-by-step reasoning, which is critical for the strong model to learn from the weak model. It enables the strong model to verify whether each step is correct or incorrect and learn how to break down the whole question into smaller steps. In contrast, the direct answer approach may mislead the model due to the lack of reasoning paths, while incorrect labels may cause more harm than using chain-of-thought, as strong model can learn nothing but false results. We conclude that for reasoning tasks within weak-to-strong generalization, chain-of-thought prompting significantly aids the learning process. Moreover, it may prove beneficial in other tasks and areas under weak-to-strong generalization.

\subsection{Is MATH 500 Precise Enough Compared to MATH 5000?}
\label{app:Additional Analysis}
As introduced in Section 2, the Performance Gap Recovered (PGR) quantifies the effectiveness of weak-to-strong generalization by comparing the performances of three models: weak model, strong ceiling model, and finetuned strong model. Our initial evaluations used a subset of 500 test samples (MATH500). Given this relatively small sample size, performance variations of up to 0.2 points per test sample were observed. This variation could be particularly significant when the performance gap between weak and strong ceiling models is small, potentially affecting the reliability of our results.

To validate our findings, we conducted additional evaluations on the untrained test set using models from the DeepSeek series. The results are presented in Table \ref{tab:math500_vs_math5000}.

\begin{table}[h]
\centering
\small
\resizebox{\columnwidth}{!}{
\begin{tabular}{lcc}
\hline
Model & MATH500 & MATH5000 \\
\hline
Weak Model & 11.4 & 9.34 \\  
Strong Ceiling & 19.6 & 20.12 \\
\hline
\multicolumn{3}{l}{\textbf{Stage I Models}} \\
\hline
Stage I-Threshold-30\% & 21.2 (119.51\%) & 19.96 (98.52\%) \\
Stage I-Threshold-40\% & 19.6 (100.00\%) & 17.58 (76.44\%) \\
Stage I-Threshold-50\% & 17.6 (75.61\%) & 16.84 (69.57\%) \\
\hline
\multicolumn{3}{l}{\textbf{Stage II Models}} \\
\hline
Stage I-30\% + Stage II-30\% & 21.4 (121.95\%) & 21.3 (110.95\%) \\
Stage I-30\% + Stage II-40\% & 21.8 (126.83\%) & 20.9 (107.24\%) \\
Stage I-30\% + Stage II-50\% & 19.4 (97.56\%) & 19.48 (94.06\%) \\ 
\hline
Stage I-40\% + Stage II-30\% & 20.4 (109.76\%) & 19.62 (95.36\%) \\
Stage I-40\% + Stage II-40\% & 19.8 (102.44\%) & 19.46 (93.88\%) \\
Stage I-40\% + Stage II-50\% & 17.4 (73.17\%) & 17.62 (76.81\%) \\
\hline
Stage I-50\% + Stage II-30\% & 20.6 (112.20\%) & 19.98 (98.70\%) \\
Stage I-50\% + Stage II-40\% & 20.6 (112.20\%) & 20.5 (103.53\%) \\
Stage I-50\% + Stage II-50\% & 19.4 (97.56\%) & 18.8 (87.76\%) \\
Stage I-50\% + Stage II-60\% & 18.6 (87.80\%) & 18.38 (83.86\%) \\
\hline
\end{tabular}
}
\caption{Performance comparison between MATH500 and MATH5000 test sets. Numbers in parentheses represent PGR values.}
\label{tab:math500_vs_math5000}
\end{table}

The results in Table \ref{tab:math500_vs_math5000} demonstrate that our framework achieves consistent performance across both MATH500 and MATH5000. While the absolute accuracy values remain similar, the slightly lower PGR on MATH5000 can be attributed to the weaker baseline performance of the weak model. However, this difference does not significantly impact our framework's effectiveness. These findings confirm that MATH500 serves as a reliable representative subset for evaluating model performance using PGR, and our framework maintains its efficacy for weak-to-strong reasoning across different evaluation scales.

\subsection{Filtering Implications on Other Datasets}
To validate the broader applicability of our framework, we conducted additional experiments on the SciQ classification task \cite{sciq} following the experimental protocol from \cite{OAIweaktostrong}. We used Qwen-1.8B as the weak supervisor and evaluated two stronger student models: Qwen-7B and Qwen-14B, employing absolute logits filtering with a threshold of 0.6. For consistency with prior work, we aligned hyperparameters with those from OpenAI’s official repository.The results are presented in Table \ref{tab:other_dataset}.

\begin{table}[h]
\resizebox{\columnwidth}{!}{
\begin{tabular}{lcc}
\hline
               & Accuracy & PGR     \\ \hline
\multicolumn{3}{l}{Qwen-7B}                                          \\ \hline
Weak Model     & 83.8            &  /             \\
Strong Ceiling & 90.0            &  /             \\
Conventional Weak-to-Strong  & 87.3             &  56.5\%            \\
Our Stage I        & 87.7           & 62.9\%(+6.4\%) \\
Our Stage II        & 87.9          & 66.1\%(+9.6\%) \\ \hline
\multicolumn{3}{l}{Qwen-14B}                                          \\ \hline
Weak Model     & 83.8            &  /             \\
Strong Ceiling & 93.5            &  /             \\
Conventional Weak-to-Strong  & 88.6             &  49.5\%            \\
Our Stage I        & 89.4           & 57.7\%(+8.2\%) \\
Our Stage II        & 89.7          & 60.8\%(+11.3\%) \\
\hline
\end{tabular}
}
\caption{Performance of our framework on the SciQ classification task with Qwen model series.}
\label{tab:other_dataset}
\end{table}

Our framework demonstrates consistent improvements across both stages, even in classification tasks distinct from mathematical reasoning. For example, Qwen-14B’s PGR improved by 8.3\% from Stage I to Stage II, while accuracy increased from 0.894 to 0.897. These results suggest that our two-stage approach effectively generalizes to diverse task formats and model scales, balancing supervision quality and question utility to mitigate overfitting. The incremental gains across stages further underline the importance of addressing both label noise and data degeneration in weak-to-strong generalization.

% \subsection{Theoretical Analysis}
% According to the expansion theory proposed by \citet{theoreticalanalysisweaktostrong}, weak-to-strong generalization emerges due to two important effects: pseudolabel correction and coverage expansion, correcting wrong labels from the weak teacher and generalizing to teacher's inconfident samples. For given student model $f$, covered subset $S$, unseen subset $T$,

\section{Additional Experimental Results}
\subsection{Detailed Analysis of Section Diversity Shifts}
\label{app:div_whole}
\begin{figure}[t]
\centering
  \includegraphics[width=0.8\columnwidth]{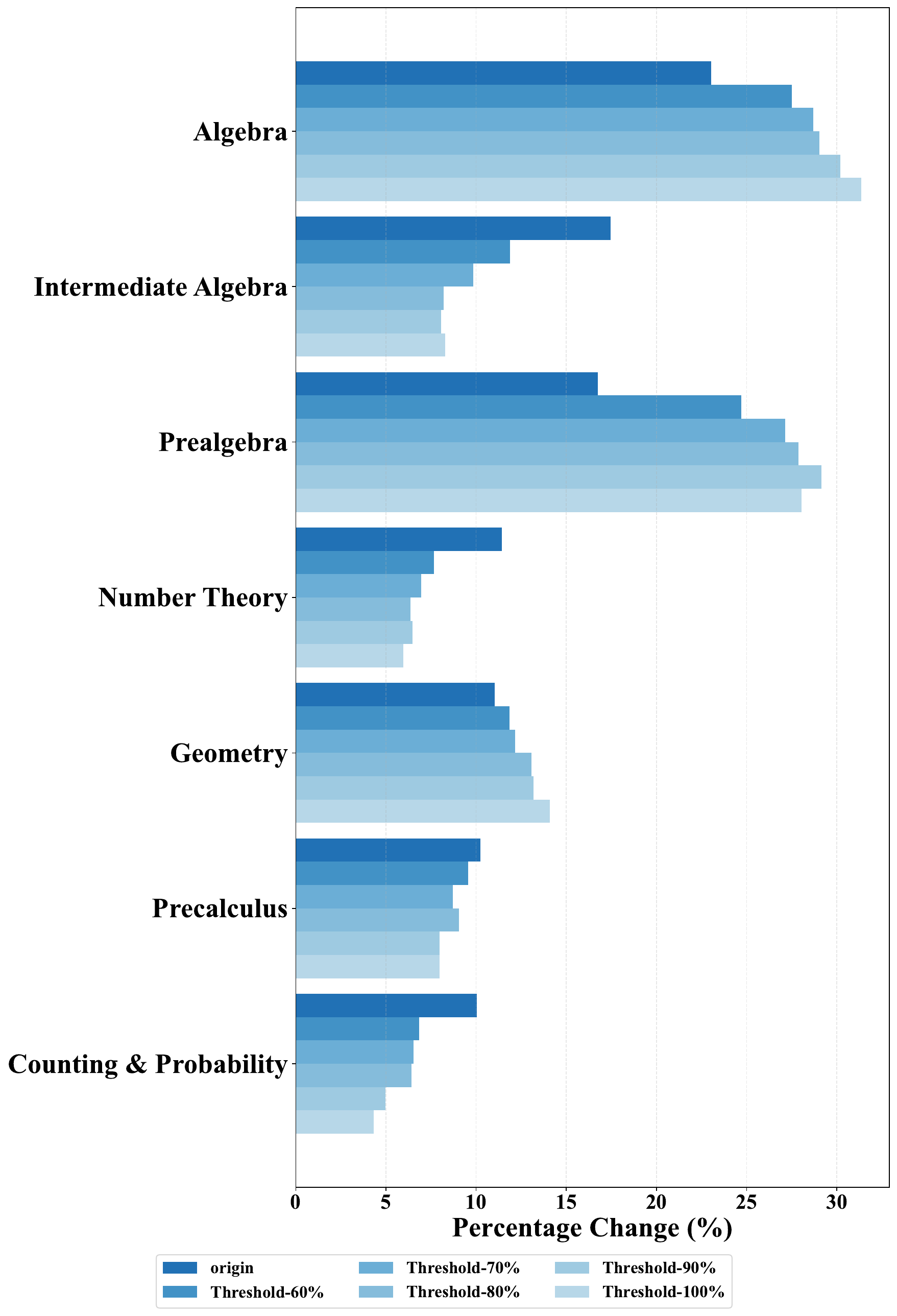}
  \caption{Changes in topic distribution across filtering thresholds for all mathematical categories in Stage I. (Llama 3 MATH) Filtering causes shifts in topic distribution, with minor categories seeing more reductions.
  % A detailed analysis of all categories is provided in Appendix \ref{appendix}.
  }
  \label{fig:div1-whole}
  \skip -0.2in
\end{figure}

\begin{figure}[!htb]
\centering
  \includegraphics[width=0.8\columnwidth]{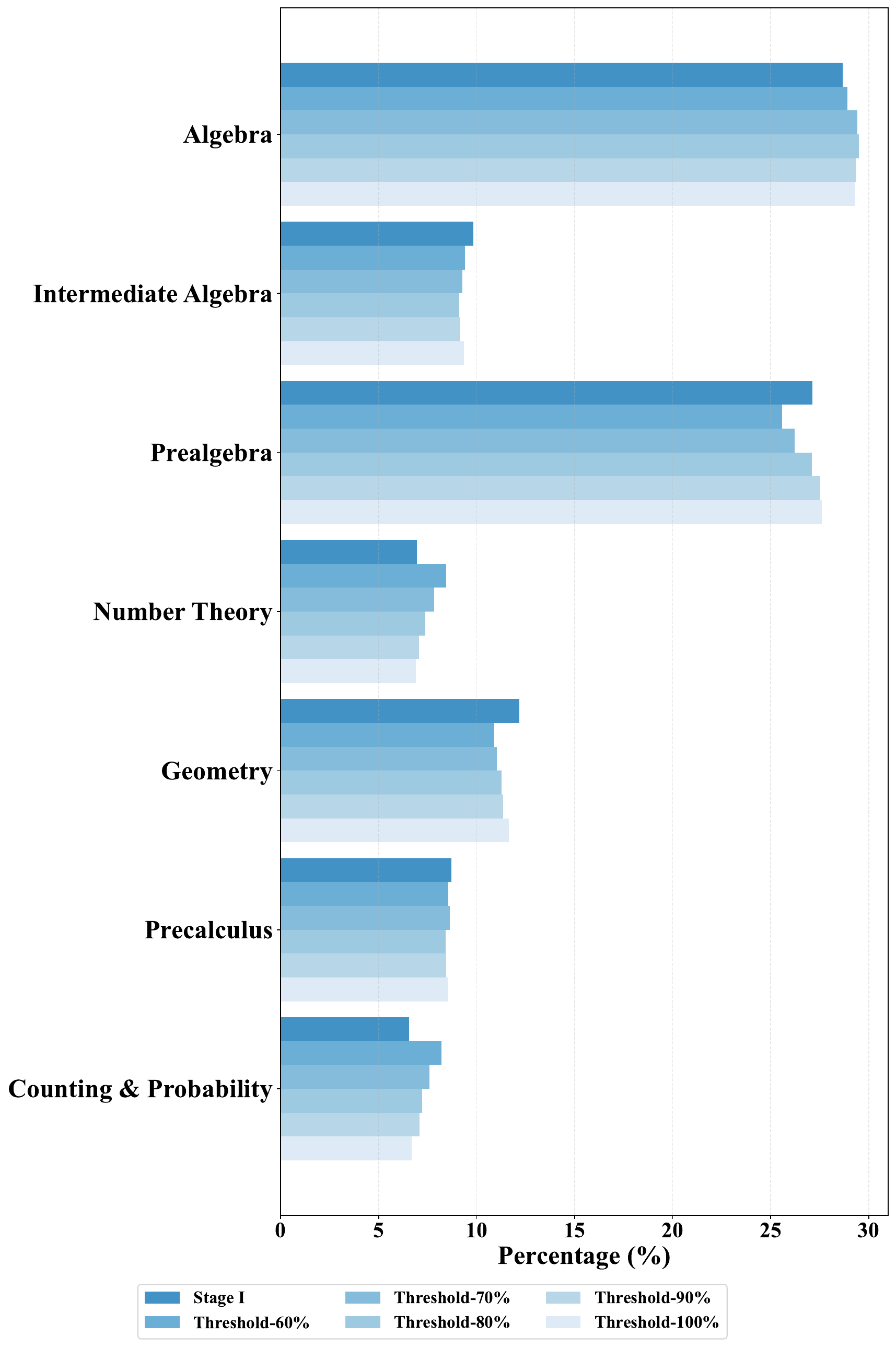}
  \caption{Changes in topic distribution across filtering thresholds for all mathematical categories in Stage II.(Llama 3 MATH Stage I-Threshold-70\%) We observe recovery in several minor categories, while sections including algebra, intermediate algebra, prealgebra are also effected by difficulty.
  % A detailed analysis of all categories is provided in Appendix \ref{appendix}.
  }
  \label{fig:div2-whole}
  \skip -0.2in
\end{figure}

In this appendix, we analyze how filtering thresholds affect section distribution in both stages of our framework. As shown in Figure \ref{fig:div1-whole} for Stage I, increasing the filtering threshold leads to a noticeable reduction in several minor categories, negatively impacting the strong model's ability to generalize effectively across a diverse range of topics. For Stage II, Figure \ref{fig:div2-whole} demonstrates how Llama 3 MATH (Stage I-Threshold-70\%) recovers some minor categories, revealing the trade-off between filtering accuracy and maintaining category diversity. We provide detailed distributions to illustrate these changes across mathematical categories.
% In this appendix, we provide a comprehensive analysis of the changes in section distribution across filtering thresholds for both Stage I and Stage II of our framework. As shown in Figure \ref{fig:div1-whole} for Stage I, increasing the filtering threshold leads to a noticeable reduction in several minor categories, negatively impacting the strong model's ability to generalize effectively across a diverse range of topics. Similarly, in Stage II, as depicted in Figure \ref{fig:div2-whole} for Llama 3 MATH (Stage I-Threshold-70\%), we observe a recovery in certain minor categories, highlighting the delicate balance between filtering for improved accuracy and maintaining the diversity necessary for robust generalization. Detailed breakdowns of these shifts are provided to offer a clearer understanding of how the filtering process influences the training data distribution across various mathematical categories.

% \clearpage

\subsection{Numeric Results of All Models and Datasets}
We present the numerical results for all models and datasets used in the experiments. It includes performance metrics for different configurations across the GSM8K and MATH benchmarks, showcasing the impact of various stages and filtering thresholds on model performance.

% \begin{figure}[h]
% \centering
%   \includegraphics[width=0.8\columnwidth]{figures/div1—whole-4.pdf}
%   \caption{Changes in topic distribution across filtering thresholds for all mathematical categories in Stage I. (Llama 3 MATH) Filtering causes shifts in topic distribution, with minor categories seeing more reductions.
%   % A detailed analysis of all categories is provided in Appendix \ref{appendix}.
%   }
%   \label{fig:div1-whole}
%   \skip -0.2in
% \end{figure}

% % \clearpage

% \begin{figure}[t]
% \centering
%   \includegraphics[width=0.8\columnwidth]{figures/div2—whole-2.pdf}
%   \caption{Changes in topic distribution across filtering thresholds for all mathematical categories in Stage II.(Llama 3 MATH Stage I-Threshold-70\%) We observe recovery in several minor categories, while sections including algebra, intermediate algebra, prealgebra are also effected by difficulty.
%   % A detailed analysis of all categories is provided in Appendix \ref{appendix}.
%   }
%   \label{fig:div2-whole}
%   \skip -0.2in
% \end{figure}

\begin{table*}[!t]
\centering
\setlength{\tabcolsep}{20pt}  % 增加列间距
\renewcommand{\arraystretch}{1.3}  % 增加行间距
\begin{tabular}{lcc}
\toprule
  &  Accuracy & Performance gap recovered(PGR) \\
\midrule
\multicolumn{3}{c}{\textbf{Basic Settings}} \\
\midrule
Weak Model & 74.8\% & 0\% \\
Strong Ceiling  & 80.36\% & 100\%\\
Conventional Weak-to-Strong & 75.2\% & 7.19\% \\
\midrule
\multicolumn{3}{c}{\textbf{Stage I}} \\
\midrule
Stage I-Threshold-30\%  & 79.37\% & 82.19\%  \\
Stage I-Threshold-40\%  & 79.51\% & 84.71\%  \\
Stage I-Threshold-50\%  & 78.99\% & 75.36\%  \\
Stage I-Threshold-60\%  & 80.07\% & 94.78\%  \\
Stage I-Threshold-70\%  & 80.28\% & 98.56\%  \\
Stage I-Threshold-80\%  & 80.06\% & 94.60\%  \\
Stage I-Threshold-90\%  & 80.13\% & 95.86\%  \\
Stage I-Threshold-100\%  & 78.16\% & 60.43\% \\
\midrule
\multicolumn{3}{c}{\textbf{Stage II based on Stage I Threshold-50\%}} \\
\midrule
Stage I-50\% + Stage II-50\% & 80.28\% & 98.56\% \\
Stage I-50\% + Stage II-60\% & 80.89\% & 109.53\% \\
Stage I-50\% + Stage II-70\% & 79.62\% & 86.69\% \\
Stage I-50\% + Stage II-80\% & 79.37\% & 82.19\% \\
\midrule
\multicolumn{3}{c}{\textbf{Stage II based on Stage I Threshold-60\%}} \\
\midrule
Stage I-60\% + Stage II-50\% & 80.28\% & 98.56\% \\
Stage I-60\% + Stage II-60\% & 81.50\% & 120.50\% \\
Stage I-60\% + Stage II-70\% & 81.04\% & 112.23\% \\
Stage I-60\% + Stage II-80\% & 81.34\% & 117.63\% \\
\midrule
\multicolumn{3}{c}{\textbf{Stage II based on Stage I Threshold-70\%}} \\
\midrule
Stage I-70\% + Stage II-60\% & 80.89\% & 109.53\% \\
Stage I-70\% + Stage II-70\% & 80.36\% & 100.00\% \\
Stage I-70\% + Stage II-80\% & 81.19\% & 114.93\% \\
Stage I-70\% + Stage II-90\% & 80.89\% & 109.53\% \\
\midrule
\multicolumn{3}{c}{\textbf{Stage II based on Stage I Threshold-80\%}} \\
\midrule
Stage I-80\% + Stage II-70\% & 80.43\% & 101.26\% \\
Stage I-80\% + Stage II-80\% & 80.33\% & 99.46\% \\
Stage I-80\% + Stage II-90\% & 80.45\% & 101.62\% \\
Stage I-80\% + Stage II-100\% & 80.74\% & 106.83\% \\
\bottomrule
\end{tabular}
\caption{Llama3 GSM8k}\label{tab:llama3 gsm8k}
\vspace{10pt}
\end{table*}

\begin{table*}[!t]
\centering
\fontsize{10.5pt}{12.5pt}\selectfont
\setlength{\tabcolsep}{20pt}
\renewcommand{\arraystretch}{1.3}
\begin{tabular}{lcc}
\toprule
 & Accuracy & Performance gap recovered(PGR) \\
\midrule
\multicolumn{3}{c}{\textbf{Basic Settings}} \\
\midrule
Weak Model & 23.8\% & 0\% \\
Strong Ceiling & 33.2\% & 100\% \\
Conventional Weak-to-Strong & 27.2\% & 36.17\% \\
\midrule
\multicolumn{3}{c}{\textbf{Stage I}} \\
\midrule
Stage I-Threshold-30\% & 27.2\% & 36.17\% \\
Stage I-Threshold-40\% & 29.8\% & 63.83\% \\
Stage I-Threshold-50\% & 30.0\% & 65.96\% \\
Stage I-Threshold-60\% & 31.4\% & 80.85\% \\
Stage I-Threshold-70\% & 34.4\% & 112.77\% \\
Stage I-Threshold-80\% & 33.2\% & 100.00\% \\
Stage I-Threshold-90\% & 32.6\% & 93.62\% \\
Stage I-Threshold-100\% & 22.6\% & -12.77\% \\
\midrule
\multicolumn{3}{c}{\textbf{Stage II based on Stage I Threshold-60\%}} \\
\midrule
Stage I-60\% + Stage II-50\% & 27.0\% & 34.04\% \\
Stage I-60\% + Stage II-60\% & 30.6\% & 72.34\% \\
Stage I-60\% + Stage II-70\% & 32.4\% & 91.49\% \\
Stage I-60\% + Stage II-80\% & 32.4\% & 91.49\% \\
Stage I-60\% + Stage II-90\% & 29.0\% & 55.32\% \\
Stage I-60\% + Stage II-100\% & 30.7\% & 73.40\% \\
\midrule
\multicolumn{3}{c}{\textbf{Stage II based on Stage I Threshold-70\%}} \\
\midrule
Stage I-70\% + Stage II-60\% & 32.2\% & 89.36\% \\
Stage I-70\% + Stage II-70\% & 32.4\% & 91.49\% \\
Stage I-70\% + Stage II-80\% & 35.2\% & 121.28\% \\
Stage I-70\% + Stage II-90\% & 34.2\% & 110.64\% \\
Stage I-70\% + Stage II-100\% & 33.2\% & 100.00\% \\
\midrule
\multicolumn{3}{c}{\textbf{Stage II based on Stage I Threshold-80\%}} \\
\midrule
Stage I-80\% + Stage II-70\% & 30.0\% & 65.96\% \\
Stage I-80\% + Stage II-80\% & 32.2\% & 89.36\% \\
Stage I-80\% + Stage II-90\% & 33.8\% & 106.38\% \\
Stage I-80\% + Stage II-100\% & 32.8\% & 95.74\% \\
\bottomrule
\end{tabular}
\caption{Llama 3 MATH}\label{tab:performance}
\vspace{10pt}
\end{table*}
\begin{table*}[!t]
\centering
\fontsize{10.5pt}{12.5pt}\selectfont  % ACL正文字体大小
\setlength{\tabcolsep}{20pt}
\renewcommand{\arraystretch}{1.3}
\begin{tabular}{lcc}
\toprule
Model & Accuracy & Performance gap recovered(PGR) \\
\midrule
\multicolumn{3}{c}{\textbf{Basic Settings}} \\
\midrule
Weak Model & 48.36\% & 0\% \\
Strong Ceiling & 75.66\% & 100\% \\
conventional Weak-to-Strong & 62.39\% & 51.39\% \\
\midrule
\multicolumn{3}{c}{\textbf{Stage I}} \\
\midrule
Stage I-Threshold-30\% & 68.68\% & 74.43\% \\
Stage I-Threshold-40\% & 70.96\% & 82.78\% \\
Stage I-Threshold-50\% & 69.74\% & 78.32\% \\
Stage I-Threshold-60\% & 70.35\% & 80.55\% \\
Stage I-Threshold-70\% & 71.11\% & 83.33\% \\
Stage I-Threshold-80\% & 69.14\% & 76.12\% \\
Stage I-Threshold-90\% & 68.38\% & 73.33\% \\
Stage I-Threshold-100\% & 67.55\% & 70.29\% \\
\midrule
\multicolumn{3}{c}{\textbf{Stage II based on Stage I Threshold-40\%}} \\
\midrule
Stage I-40\% + Stage II-30\% & 72.63\% & 88.90\% \\
Stage I-40\% + Stage II-40\% & 72.32\% & 87.77\% \\
Stage I-40\% + Stage II-50\% & 70.58\% & 81.39\% \\
Stage I-40\% + Stage II-60\% & 72.17\% & 87.22\% \\
\midrule
\multicolumn{3}{c}{\textbf{Stage II based on Stage I Threshold-60\%}} \\
\midrule
Stage I-60\% + Stage II-60\% & 70.28\% & 80.29\% \\
Stage I-60\% + Stage II-70\% & 71.49\% & 84.73\% \\
Stage I-60\% + Stage II-80\% & 70.28\% & 80.29\% \\
Stage I-60\% + Stage II-90\% & 70.28\% & 80.29\% \\
\midrule
\multicolumn{3}{c}{\textbf{Stage II based on Stage I Threshold-70\%}} \\
\midrule
Stage I-70\% + Stage II-60\% & 72.40\% & 88.06\% \\
Stage I-70\% + Stage II-70\% & 72.94\% & 90.04\% \\
Stage I-70\% + Stage II-80\% & 71.64\% & 85.27\% \\
Stage I-70\% + Stage II-90\% & 72.55\% & 88.61\% \\
\midrule
\multicolumn{3}{c}{\textbf{Stage II based on Stage I Threshold-80\%}} \\
\midrule
Stage I-80\% + Stage II-70\% & 70.20\% & 80.00\% \\
Stage I-80\% + Stage II-80\% & 70.50\% & 81.10\% \\
Stage I-80\% + Stage II-90\% & 71.47\% & 84.65\% \\
Stage I-80\% + Stage II-100\% & 70.35\% & 80.55\% \\
\bottomrule
\end{tabular}
\caption{Deepseek-GSM8K}\label{tab:llama3 math}
\vspace{10pt}
\end{table*}

% \subsection{Deepseek-GSM8K}

% \subsection{Deepseek-MATH}
\begin{table*}[!t]
\centering
\fontsize{10.5pt}{12.5pt}\selectfont
\setlength{\tabcolsep}{20pt}
\renewcommand{\arraystretch}{1.3}
\begin{tabular}{lcc}
\toprule
Model & Accuracy & Performance gap recovered(PGR) \\
\midrule
\multicolumn{3}{c}{\textbf{Basic Settings}} \\
\midrule
Weak Model & 11.4\% & 0\% \\
Strong Ceiling & 19.6\% & 100\% \\
conventional Weak-to-Strong & 16.8\% & 65.85\% \\
\midrule
\multicolumn{3}{c}{\textbf{Stage I}} \\
\midrule
Stage I-Threshold-30\% & 21.2\% & 119.51\% \\
Stage I-Threshold-40\% & 19.6\% & 100.00\% \\
Stage I-Threshold-50\% & 17.6\% & 75.61\% \\
Stage I-Threshold-60\% & 15.8\% & 53.66\% \\
Stage I-Threshold-70\% & 16.4\% & 60.98\% \\
Stage I-Threshold-80\% & 15.0\% & 43.90\% \\
Stage I-Threshold-90\% & 12.0\% & 7.32\% \\
\midrule
\multicolumn{3}{c}{\textbf{Stage II based on Threshold-30\%}} \\
\midrule
Stage I-30\% + Stage II-30\% & 21.4\% & 121.95\% \\
Stage I-30\% + Stage II-40\% & 21.8\% & 126.83\% \\
Stage I-30\% + Stage II-50\% & 19.4\% & 97.56\% \\
Stage I-30\% + Stage II-60\% & 19.2\% & 95.12\% \\
Stage I-30\% + Stage II-70\% & 19.0\% & 92.68\% \\
\midrule
\multicolumn{3}{c}{\textbf{Stage II based on Threshold-40\%}} \\
\midrule
Stage I-40\% + Stage II-30\% & 20.4\% & 109.76\% \\
Stage I-40\% + Stage II-40\% & 19.8\% & 102.44\% \\
Stage I-40\% + Stage II-50\% & 17.4\% & 73.17\% \\
Stage I-40\% + Stage II-60\% & 18.0\% & 80.49\% \\
\midrule
\multicolumn{3}{c}{\textbf{Stage II based on Threshold-50\%}} \\
\midrule
Stage I-50\% + Stage II-30\% & 20.6\% & 112.20\% \\
Stage I-50\% + Stage II-40\% & 20.6\% & 112.20\% \\
Stage I-50\% + Stage II-50\% & 19.4\% & 97.56\% \\
Stage I-50\% + Stage II-60\% & 18.6\% & 87.80\% \\
\bottomrule
\end{tabular}
\caption{Deepseek-MATH}\label{tab:results}
\vspace{10pt}
\end{table*}

\end{document}